\documentclass[10pt]{article} 
\usepackage{omnipiano_preprint}

\usepackage{amsmath,amsfonts,bm}

\def\eqref#1{equation~\ref{#1}}

\def\1{\bm{1}}

\DeclareMathAlphabet{\mathsfit}{\encodingdefault}{\sfdefault}{m}{sl}
\SetMathAlphabet{\mathsfit}{bold}{\encodingdefault}{\sfdefault}{bx}{n}

\usepackage{hyperref}
\hypersetup{colorlinks=true,linkcolor=dark-blue,citecolor=dark-blue,urlcolor=dark-blue,
  pdftitle={OmniPiano: Diverse Dexterous Piano-Playing Challenges for Standard, Robust, Safe, and Multi-Agent RL},
  pdfauthor={Fan Xu, Fan Yang, Yifan Xu, Ruiyuan Zhang, Yimou Wu, Nuo Yi, Muning Wen, Guang Chen, Costas Spanos, Shangding Gu}}
\usepackage{url}
\usepackage{booktabs}
\usepackage{tabularx}
\usepackage{float}
\usepackage{graphicx}
\usepackage{amssymb}
\usepackage{subcaption}
\usepackage{enumitem}
\usepackage{subcaption}
\usepackage[T1]{fontenc}
\usepackage{pifont}
\usepackage{wrapfig}
\usepackage{caption}
\usepackage{array}
\usepackage{tcolorbox}
\usepackage{multirow}
\usepackage{placeins}

\usepackage{bm}

\newcolumntype{Y}{>{\raggedright\arraybackslash}X}

\newcommand{\cmark}{\ding{51}}
\newcommand{\xmark}{\ding{55}}

\usepackage{xspace}
\newcommand{\ourbenchmark}{\textbf{\emph{OmniPiano}}\xspace}

\usepackage{titlesec}
\usepackage{omnipiano_layout}

\usepackage{enumitem}

\title{OmniPiano: Diverse Dexterous Piano-Playing Challenges for Standard, Robust, Safe, and Multi-Agent RL}

\author{%
\begin{minipage}[t]{\dimexpr\textwidth-2\tabcolsep\relax}
\centering
\normalfont\normalsize
\setlength{\parskip}{0pt}
\setlength{\parindent}{0pt}
\textbf{Fan Xu\textsuperscript{1,*,$\dagger$},\;
Fan Yang\textsuperscript{2,*},\;
Yifan Xu\textsuperscript{3,*},\;
Ruiyuan Zhang\textsuperscript{4,*}}\\[1pt]
\textbf{Yimou Wu\textsuperscript{5},\;
Nuo Yi\textsuperscript{6},\;
Muning Wen\textsuperscript{2},\;
Guang Chen\textsuperscript{7}}\\[1pt]
\textbf{Costas Spanos\textsuperscript{1},\;
Shangding Gu\textsuperscript{1,2,$\ddagger$}}\\[3pt]
\normalfont\small
\textsuperscript{1}University of California, Berkeley\quad
\textsuperscript{2}Shanghai Jiao Tong University\\[1pt]
\textsuperscript{3}The Chinese University of Hong Kong, Shenzhen\\[1pt]
\textsuperscript{4}Japan Advanced Institute of Science and Technology\\[1pt]
\textsuperscript{5}The Chinese University of Hong Kong\quad
\textsuperscript{6}Nanjing University\quad
\textsuperscript{7}Tongji University\\[3pt]
\textsuperscript{*}Equal contribution\quad
\textsuperscript{$\dagger$}Project Lead\quad
\textsuperscript{$\ddagger$}Corresponding author
\end{minipage}%
}

\begin{document}

\maketitle

\fancyhead{}

\renewcommand{\headrulewidth}{0pt}

\begin{abstract}

Dexterous robotic manipulation remains a major challenge in robotics, exemplified by piano playing, which requires coordinated control of fingers and provides a demanding testbed for reinforcement learning (RL). Although previous work established a benchmark for robotic piano playing, it is restricted to two hands and does not support unified evaluation.
To address this gap, we introduce \ourbenchmark, a benchmark that provides one to up to five Shadow Hands and supports standard, robust, safe, and multi-agent RL within a shared piano-playing task family. \ourbenchmark is developed for scalable and diverse piano-playing tasks with configurable perturbations, explicit safety constraints, and decentralized cooperation settings. Particularly, to facilitate usability and extendability, \ourbenchmark adopts a highly modular design and provides comprehensive tasks to support RL study of task performance, robustness, safety, and cooperation in dexterous control. With at least 912 task settings and 36 baseline algorithms, \ourbenchmark further incorporates LLM-based agents to broaden the evaluation scope and reveal new insights. Extensive evaluations show that state-of-the-art RL algorithms and frontier LLM-based agents still struggle on  challenging piano-playing tasks, even under standard RL settings, highlighting substantial room for improvement in dexterous control. The code, dataset, tutorial, and demo are available at \url{https://sail-research-lab.github.io/omnipiano/}.

\end{abstract}

\section{Introduction} 
\label{sec:introduction}
Reinforcement learning (RL) has made substantial progress in games \citep{mnih2015human,vinyals2019grandmaster}, continuous control \citep{duan2016benchmarking}, and robotic manipulation \citep{levine2016end}, yet learning precise, coordinated behavior in high-dimensional, contact-rich systems remains challenging. Dexterous piano playing exemplifies this challenge and provides a suitable testbed for evaluating RL methods under precise temporal, spatial and coordination requirements. In this task, RL methods learn to coordinate finger movements to press, hold, and release keys at the times specified by a Musical Instrument Digital
Interface (MIDI) score \citep{xu2022learningplaypianodexterous}, which provides a precise temporal target for evaluating RL performance through key-activation accuracy and timing.

The representative benchmark, RoboPianist, uses simulated Shadow Hands for dexterous piano playing \citep{zakka2023robopianist}. However, its limited two-hand setting hinders scalability as the number of hands and control dimensions increase. More broadly, a unified dexterous piano playing benchmark for robust RL, safe RL, and multi-agent RL is still lacking. These gaps motivate a benchmark that extends the previous work to multi-hand configurations and supports systematic evaluation under perturbations, explicit safety constraints, and decentralized control, with shared musical objectives and performance measures.

\begin{figure}[H]
    \centering
    \includegraphics[width=0.9\linewidth]{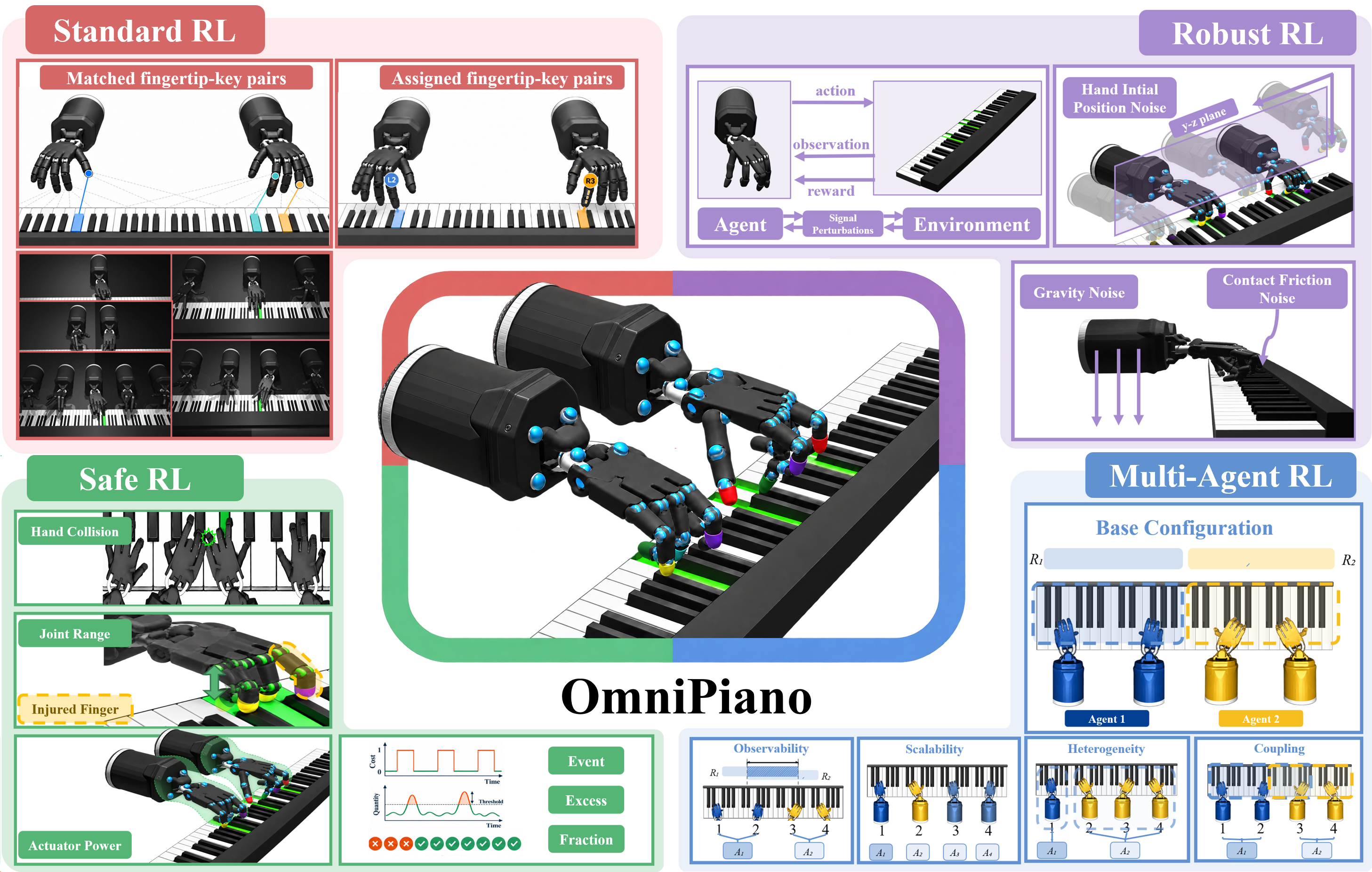}
    \caption{The overview of \ourbenchmark. A unified benchmark for learning dexterous piano playing. }
    \label{fig:framework}
\end{figure}

To address this gap, we introduce \ourbenchmark, a benchmark for standard, robust, safe, and multi-agent RL through dexterous piano playing. As illustrated in Figure~\ref{fig:framework}, \ourbenchmark comprises four tracks built on a shared piano-playing core, each varying a different aspect of the same task: \emph{standard RL} extends the number of hands under two kinds of reward formulations, \emph{robust RL} injects interaction and environment perturbations, \emph{safe RL} introduces safety-cost constraints alongside task rewards, and \emph{multi-agent RL} distributes the hands across decentralized agents. Rather than combining every factor within a single environment, this design preserves the core piano-playing task while varying one factor at a time, so that performance differences can be attributed to a single design choice. As shown in Table~\ref{tab:related_work_comparison}, \ourbenchmark leverages challenging dexterous control with broad coverage of RL evaluation dimensions. Its unified task setting supports robust, safe, and multi-agent RL alongside LLM-agent evaluation, bringing together capabilities that remain fragmented across existing benchmarks.

\ourbenchmark is implemented in MuJoCo and exposes all four task families through consistent Gymnasium (single-agent) and PettingZoo (multi-agent) interfaces \citep{todorov2012mujoco,terry2021pettingzoo,kwiatkowski2024gymnasium}, enabling diverse RL algorithms to be compared across settings. The benchmark comprises 72, 216, 480, and 144 task settings across the standard, robust, safe, and multi-agent tracks, respectively. Our evaluation covers 10 standard RL methods, 4 robust RL methods, 14 safe RL methods, and 8 multi-agent RL methods, together with 4 LLM-based agents, for a total of 912 task settings and 36 baselines (as shown in Appendix Table~\ref{tab:supported_baselines}).
Overall, these tracks provide a foundation for studying RL in high-dimensional dexterous control.

Our main contributions are as follows:

\begin{itemize}[leftmargin=*]

    \item \textbf{A unified evaluation framework beyond single evaluation track.}
    We extend the piano-playing task into four task families covering standard, robust, safe, and multi-agent RL, with at least 912 tasks in total.

    \item \textbf{A comprehensive and reproducible foundation for algorithm comparison.}
    We provide 36 representative RL baselines and 4 LLM-based agents, along with task-specific metrics, standardized evaluation protocols, and unified Gymnasium and PettingZoo interfaces, enabling systematic and reproducible comparison across the benchmark.

    \item \textbf{A scalable benchmark for multi-hand dexterous control.}
    We introduce \ourbenchmark, providing one to up to five Shadow Hands and at most 111 continuous action dimensions. This expansion enables systematic investigation of how RL scales with control dimensionality, workspace coupling, and coordination demands.

\end{itemize}

\begin{table}[H]
\caption{The comparison of \ourbenchmark with related reinforcement-learning benchmarks.}
\label{tab:related_work_comparison}
\centering
\small
\setlength{\tabcolsep}{3.5pt}
\renewcommand{\arraystretch}{1.08}

\begin{tabularx}{\linewidth}{@{}Xcccc@{}}
\toprule
Feature
& RoboPianist
& Robust-Gym.
& Safety-Gym.
& \textbf{\ourbenchmark} \\
\midrule

Action dimension
& 45
& 1--30
& 2--17
& \textbf{23--111} \\

Dexterous piano playing
& \cmark
& \xmark
& \xmark
& \textbf{\cmark} \\

Robust RL track
& \xmark
& \cmark
& \xmark
& \textbf{\cmark} \\

Safe RL track
& \xmark
& \cmark
& \cmark
& \textbf{\cmark} \\

Multi-agent RL track
& \xmark
& \cmark
& \cmark
& \textbf{\cmark} \\

Unified task setting
& \xmark
& \xmark
& \xmark
& \textbf{\cmark} \\

LLM-agent evaluation
& \xmark
& \xmark
& \xmark
& \textbf{\cmark} \\

\bottomrule
\end{tabularx}
\vspace{-2mm}
\end{table}

Due to space limits, we provide preliminaries in Appendix~\ref{app:background_problem} and related work in Appendix~\ref{app:related_work}.

\section{OmniPiano}
\label{sec:benchmark}

We now introduce \ourbenchmark, a unified benchmark for standard, robust, safe, and multi-agent reinforcement learning through dexterous piano playing. Each benchmark task comprises three components: an embodied agent configuration comprising one to five Shadow Hands, a musical objective specified by a MIDI score, and a task-specific design that modifies the agent--environment interaction. This modular construction yields four complementary settings: standard RL measures algorithm performance on standard tasks; robust RL introduces controlled perturbations to observations, actions, rewards, and the physical environment; safe RL augments the task with explicit costs for joint, collision, power, and injury constraints; and multi-agent RL distributes the hands across decentralized agents that pursue a shared musical objective. In the following sections, we describe the task and environment foundations, with the design of each benchmark setting.

\subsection{Task and Environment Foundations}
\label{sec:task_environment}
In this section, we describe the shared task and environment foundations of \ourbenchmark. We use the RoboPianist~\citep{zakka2023robopianist} framework as our starting point, built on the MuJoCo physics engine~\citep{todorov2012mujoco, Tunyasuvunakool_2020}. Based on this framework, we introduce a new piano-playing environment that supports multiple Shadow Hands and diverse task designs. The environment provides a unified interface for users to configure the number of hands, songs to play, and task designs for standard, robust, safe, and multi-agent reinforcement learning. And the environment is compatible with popular reinforcement learning libraries such as Stable-Baselines3~\citep{stable-baselines3}, RLlib~\citep{liang2018rllibabstractionsdistributedreinforcement}, OmniSafe~\citep{ji2024omnisafe} and torchRL~\citep{bou2023torchrldatadrivendecisionmakinglibrary} enabling users to easily train and evaluate their policies.

\subsection{Standard RL Benchmark Design}
\label{sec:standard_design}

\begin{description}[
    leftmargin=0pt,
    labelindent=0pt,
    itemsep=1pt,
    parsep=0pt,
    topsep=2pt,
    labelsep=0.5em
]
    \item[\textbf{Hand settings.}] \ourbenchmark includes up to 150 songs and eight hand settings, with the latter varying in both the number of hands (from one to five) and their mobility configurations. Specifically, \textit{unrestricted} denotes the fully mobile setting, in which each hand can move freely across the keyboard, whereas \textit{restricted} assigns each hand to a fixed region. We introduce restricted configurations to improve the learnability of multi-hand tasks: because with the increasing of the number of hands, full mobility enlarges the joint action and coordination space, making exploration and credit assignment increasingly difficult for reinforcement learning.

    \item[\textbf{Reward design.}] Following RoboPianist, we formulate the reward as Equation~\ref{eq:ot reward}:
    \begin{equation}
    \label{eq:ot reward}
        r_t =
        r_t^{\mathrm{key}}
        +
        r_t^{\mathrm{match}}
        -
        \lambda_{\mathrm{energy}} r_t^{\mathrm{energy}},
    \end{equation} where $r_t^{\mathrm{key}}$ rewards pressing the target keys while penalizing false-positive key activations, $r_t^{\mathrm{match}}$ encourages assigned fingertips to approach their target keys, and $r_t^{\mathrm{energy}}$ regularizes excessive joint actuation. We provide two variants that differ only in the fingertip-to-key assignment, \textit{fingering-annotated reward} and  \textit{optimal transport (OT) reward}. Details can be seen in \ref{app:detailed_reward_setting}. And evaluation metric F1 is described in Appendix \ref{app:evaluation_metric}.
\end{description}

\begin{figure}[H]
    \centering
    \includegraphics[width=\linewidth]{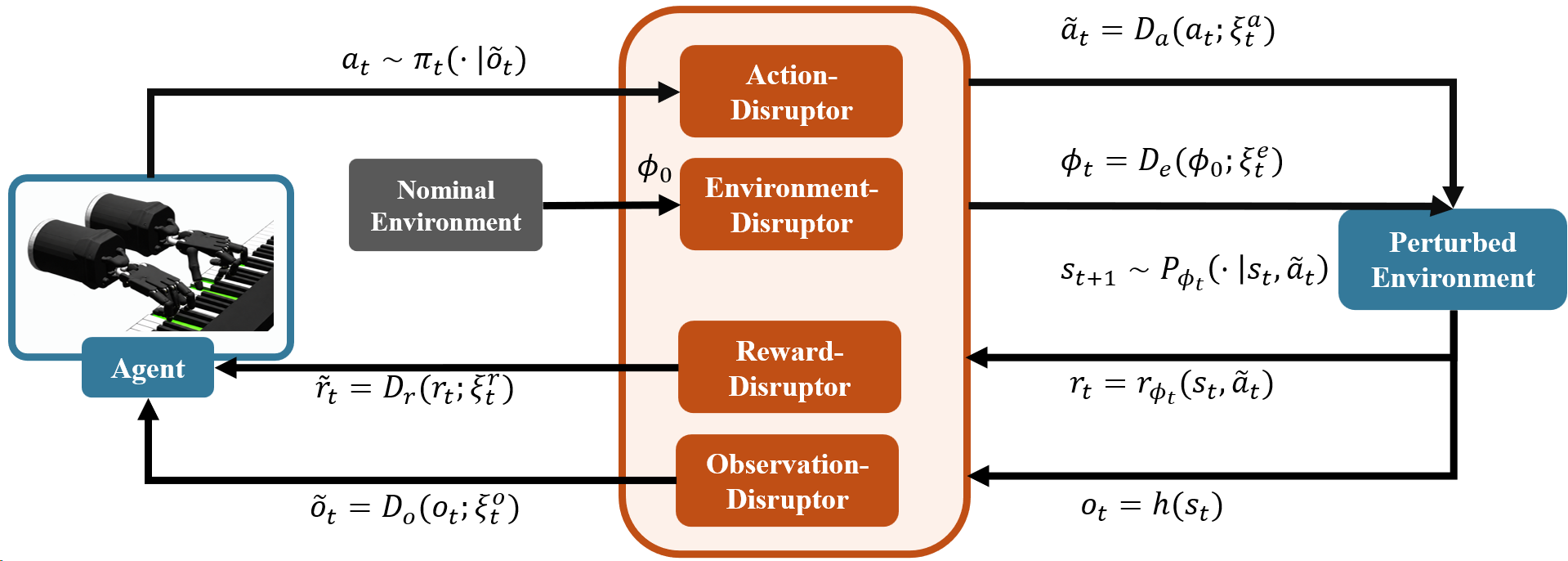}
    \caption{The overview of the Robust RL task design. Robustness perturbations are introduced through four independent disruptors acting on the agent--environment interaction loop. }
    \label{fig:robustRL}
\end{figure}

\subsection{Robust RL Benchmark Design}
\label{sec:robust_design}

\ourbenchmark evaluates robustness in the closed-loop agent--environment interaction. Illustrated in Figure~\ref{fig:robustRL}, we consider two types of perturbation: \emph{(i) signal perturbations}, which affect the observations, actions, or rewards exchanged during interaction; and \emph{(ii) environment perturbations}, which alter the simulated physical conditions, including gravity, fingertip--key contact-friction, and the initial pose of each hand. Table~\ref{tab:robustness_design} summarizes the resulting perturbation axes and canonical levels.

\begin{description}[
    leftmargin=0pt,
    labelindent=0pt,
    itemsep=1pt,
    parsep=0pt,
    topsep=2pt,
    labelsep=0.5em
]
    \item[\textbf{Signal perturbations.}] The operators $D_o$, $D_a$, and $D_r$ perturb observations, actions, and rewards, respectively. $D_o$ perturbs the observed hand and piano states, while keeping the fingering targets unchanged; $D_a$ models deviations between commanded and executed actions; and $D_r$ perturbs the reward signal.

    \item[\textbf{Environment perturbations.}] The operator $D_e$ modifies environment through gravity, contact-friction, and initial hand-pose perturbations, affecting fingertip--key interaction. Gravity and friction can vary per episode or step, while hand-pose offsets are sampled at reset.

    \item[\textbf{Composable perturbations.}] All perturbation channels can be combined flexibly, with independently specified distributions, magnitudes, and frequencies, enabling controlled evaluation of both isolated and compound robustness effects under the same underlying musical objective.
\end{description}

\begin{figure}[h]
    \centering
    \includegraphics[width=\linewidth]{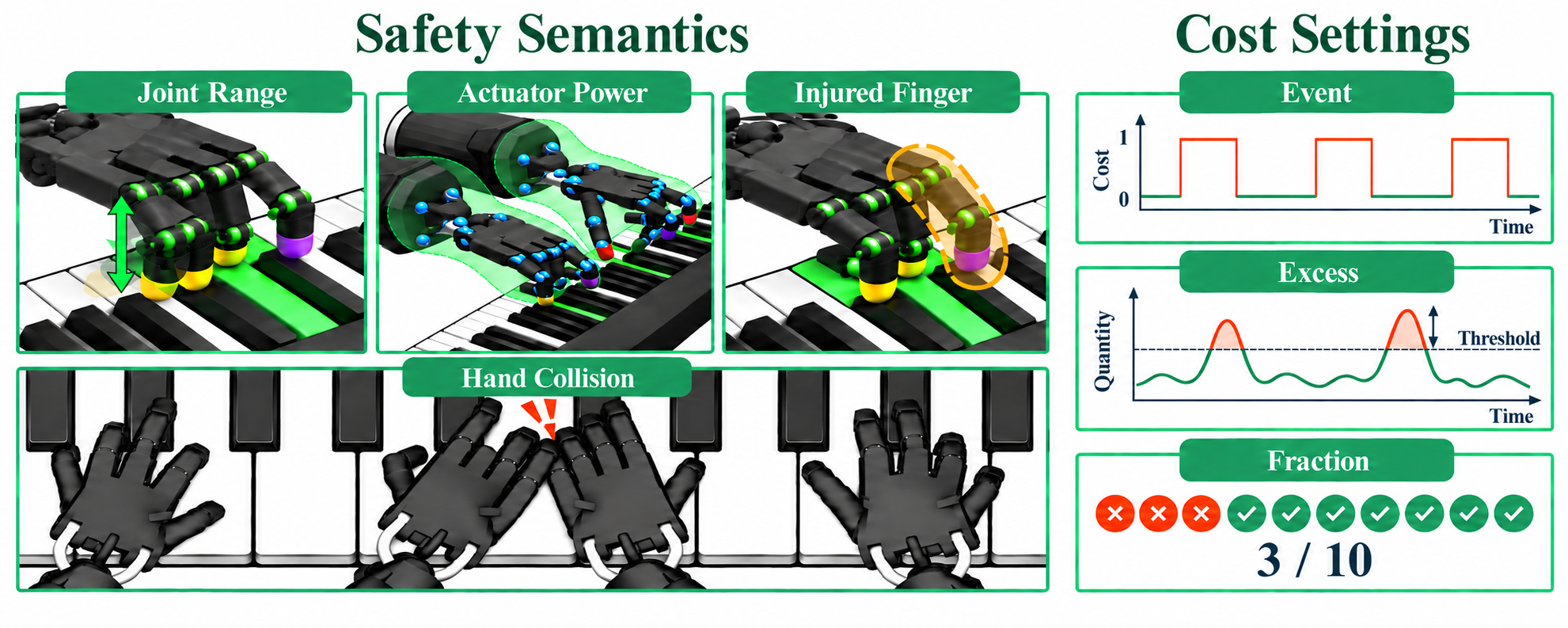}
    \caption{The overview of the safety task design in \ourbenchmark. Four safety semantics, Joint Range, Actuator Power, Injured Finger, and Hand Collision, define the constrained behaviors. Event, Fraction, and Excess provide alternative measures of constraint violations.}
    \label{fig:safety_design}
\end{figure}

\subsection{Safe RL Benchmark Design}
\label{subsec:safety_design}

Accurate piano playing does not imply safe control. Policies with similar musical performance may differ substantially in their joint configurations, contact forces, and actuation demands. \ourbenchmark therefore introduces safety cost constraints to enable systematic study of musical performance
under explicit safety constraints and different safety-performance trade-offs.

\begin{description}[
    leftmargin=0pt,
    labelindent=0pt,
    itemsep=1pt,
    parsep=0pt,
    topsep=2pt,
    labelsep=0.5em
]
    \item[\textbf{Reward design.}] All safety tasks share a common musical objective and use a fingering-free OT reward as Equation~\ref{eq:ot reward}. By matching fingertips to target keys without prescribing a fixed fingering, this reward allows policies to adapt finger assignments to the safety constraint. We remove the actuator-power penalty from the base reward so that power is handled through the independent safety cost, making the trade-off between musical performance and constraint satisfaction explicit.

    \item[\textbf{Safety semantics.}] As illustrated in Figure~\ref{fig:safety_design}, \ourbenchmark considers four safety semantics:
    (1) \textbf{Joint Range} restricts selected joint movements to prescribed intervals;
    (2) \textbf{Actuator Power} limits instantaneous power;
    (3) \textbf{Injured Finger} applies a power limit to each actuator controlling a designated thumb; this constraint represents restricted use of an injured finger; and
    (4) \textbf{Hand Collision} limits unintended contact forces between hands in the shared keyboard workspace.
    These constraints capture different physical safety aspects, with thresholds chosen as benchmark parameters for controlled evaluation.
    
    \item[\textbf{Cost settings.}] Violations are quantified through three cost formulations:
    (1) \textbf{Event} records whether a violation occurs at the current control step;
    (2) \textbf{Fraction} measures the proportion of violations at each control step; and
    (3) \textbf{Excess} measures the normalized deviation beyond an admissible interval or threshold.
    They capture violation occurrence, proportion, and severity, respectively. The applicable formulation depends on the design of the safety signal. For example, Joint Range combined with Fraction measures the proportion of selected joints outside their prescribed ranges.

    \item[\textbf{Task construction and evaluation.}] Each task comprises a musical piece and hand configuration with the shared OT reward, one selected safety semantic, one selected cost setting, and an episodic cost limit. Policies maximize musical return subject to this limit, and evaluation jointly assesses playing quality and constraint satisfaction.
\end{description}

\begin{figure}[h]
    \centering
    \includegraphics[width=1\linewidth]{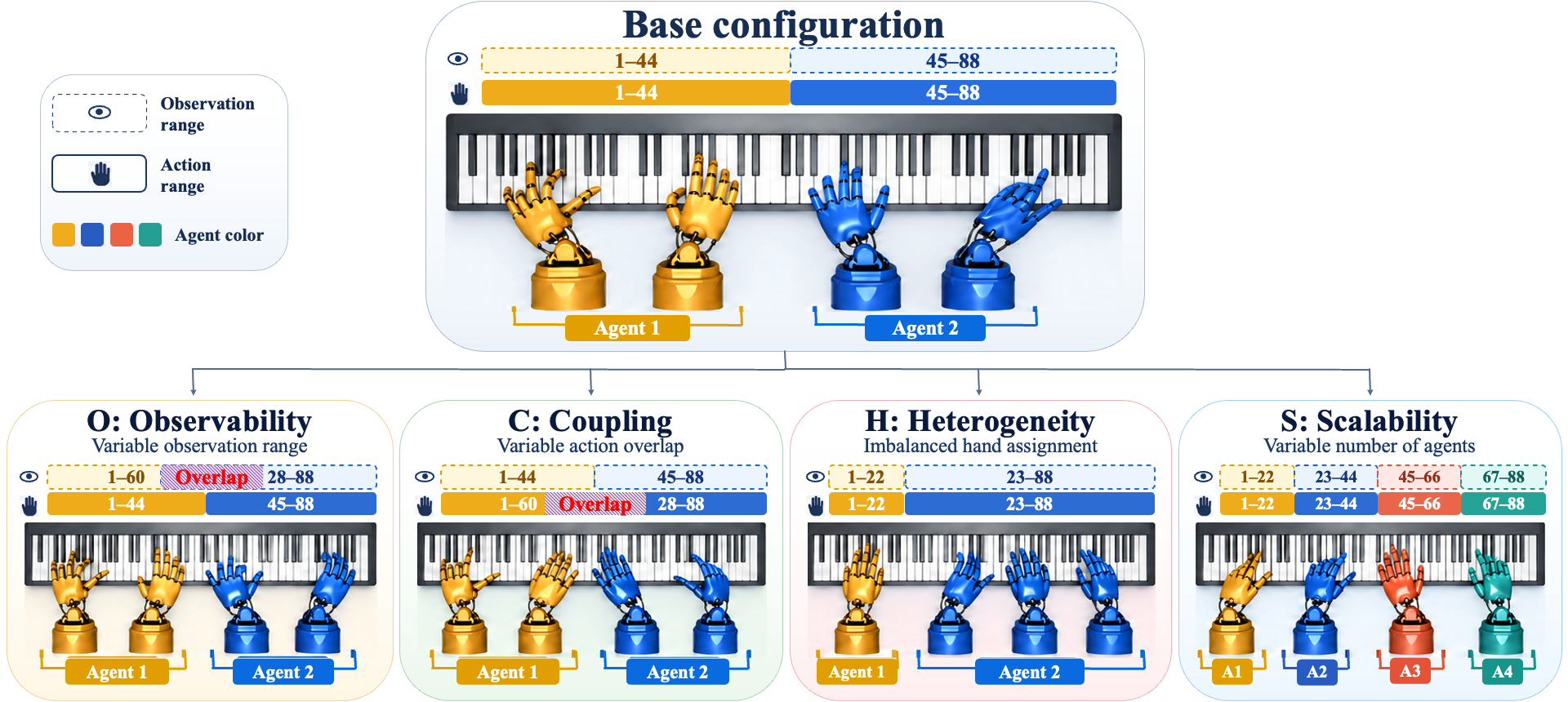}
    \caption{The overview of the multi-agent RL task design. The configuration shown is a typical example. In practice, users can freely configure the MIDI score, number of hands and agents, hand-to-agent assignments, and agent-specific observation and action ranges and overlaps.}
    \label{fig:marl_illustration}
\end{figure}
\subsection{Multi-Agent RL Benchmark Design}
\label{sec:multiagent_design}

\ourbenchmark organizes MARL evaluation around four configurable dimensions, \textbf{Scalability, Coupling, Heterogeneity}, and \textbf{Observability (SCHO)}. Under the same hand configuration, musical objective, and shared reward, these dimensions vary different aspects of multi-agent control setup. Specifically, \textbf{Scalability} varies how many agents control the same set of Shadow Hands. More agents will split the same physical task across more independent decision makers. \textbf{Coupling} varies the overlap between agents' keyboard action ranges. More overlap lets multiple agents act on more same keys, increasing coordination demands and the chance of interfering actions. \textbf{Heterogeneity} varies how the Shadow Hands are divided among agents. Balanced assignments give agents similar control loads, while imbalanced assignments create unequal action dimensions and keyboard responsibilities. Finally, \textbf{Observability} varies the scope of each agent's local keyboard and teammate observations. Narrower observations give agents less information about the shared task without changing what they control. An illustrative 4-hands example is provided as Figure~\ref{fig:marl_illustration}. 

\section{Experiments and Analysis}
\label{sec:experiments}

We will first introduce the task settings, evaluation results and key findings for standard RL, robust RL, safe RL, and multi-agent RL in Sections~\ref{sec:standard_rl_results}, \ref{sec:robust_rl_results}, \ref{sec:safe_rl_results} and \ref{sec:marl_results}. Due to space constraints, additional experiments and detailed settings are provided in Appendix~\ref{app:additional_experiments} and Appendix~\ref{app:experimental_settings}, respectively.

\begin{figure}[H]
    \centering
    \includegraphics[width=\linewidth]{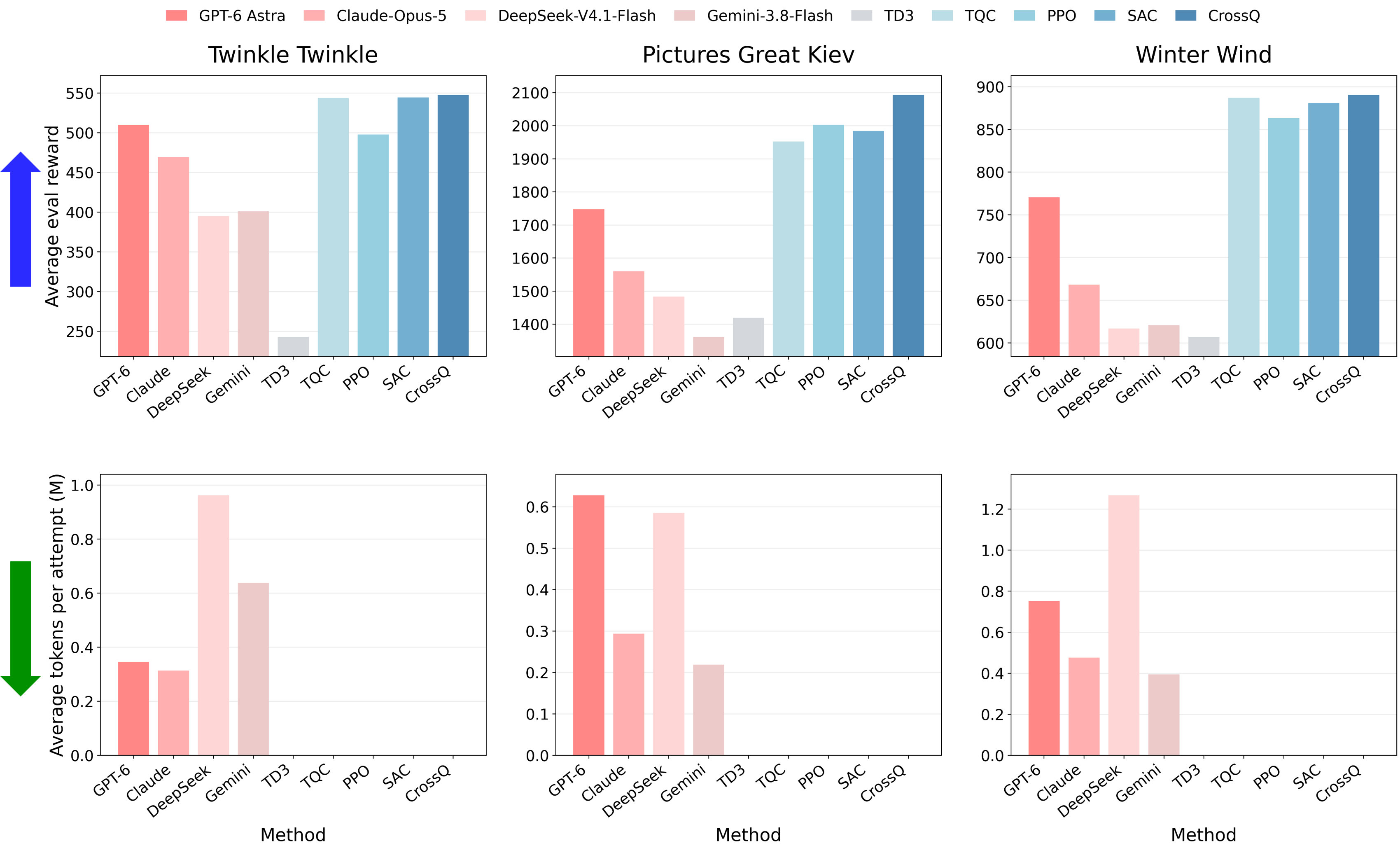}
    \caption{Average evaluation \textbf{reward}  (\textcolor{blue}{$\bm{\uparrow}$}: top row, higher is better) and mean \textbf{token usage} (\textcolor{green!60!black}{$\bm{\downarrow}$}: bottom row, lower is better) per optimization attempt for LLMs across ten keyframe-optimization iterations and RL methods over 5M training steps in the two-hand setting on three piano pieces.}
    \label{fig:twohand_mean_reward_tokens}
\end{figure}

\subsection{Evaluation of Standard RL Baselines}
\label{sec:standard_rl_results}
We first examine the relationship between song difficulty and task performance across representative piano pieces (Table~\ref{tab:song_difficulty_f1}). We compare episode length, measured in control steps at $20\,\mathrm{Hz}$, with note density, measured by Notes Per Second (NPS). Episode length alone does not reflect difficulty consistently: \textit{Gymnopédie No.1} is longer than \textit{French Suite No.1 Allemande} (657 vs.\ 517 steps) yet achieves higher F1 (0.88 vs.\ 0.59), while its lower NPS (1.77 vs.\ 10.89) better matches the performance gap, suggesting that NPS is more informative of task difficulty than duration alone.

We then evaluate PPO~\citep{schulman2017proximalpolicyoptimizationalgorithms} and SAC~\citep{haarnoja2018softactorcriticoffpolicymaximum} as standard RL baselines and compare them with large language models (LLMs). Experiments are conducted on three pieces of increasing difficulty: \textit{Twinkle Twinkle} (161 steps), \textit{Pictures of Great Kiev} (720 steps), and \textit{WinterWind} (314 steps), with one to five hands. For three- to five-hand tasks, both unrestricted and restricted hand assignments are considered (Figure~\ref{fig:general-llm-f1-radar}).  Additional details on the LLM-based agents are provided in Appendix~\ref{app:detailed_llm}. Across these settings, increasing the number of hands does not consistently improve performance. Meanwhile, LLM-based agents remain competitive with RL baselines and achieve higher F1 on challenging pieces (\textit{WinterWind}) in the five-hand setting. Figure~\ref{fig:general-rl-restricted-learning} further compares the two assignment schemes under standard RL. Restricted assignments achieve faster reward convergence and higher final returns than unrestricted assignments.

For the two-hand setting, we further compare PPO, SAC, CrossQ~\citep{bhatt2024crossqbatchnormalizationdeep}, TD3~\citep{10980556}, TQC~\citep{kuznetsov2022automatingcontroloverestimationbias}, and LLM-based agent methods (Figure~\ref{fig:twohand_mean_reward_tokens}, Figure~\ref{fig:twohand_f1}). Over 5M environment steps, RL methods exploit the reward signal more consistently and achieve more stable F1 improvements (Figure~\ref{fig:twohand_final_eval_reward}). In contrast, LLM-based agents can obtain F1 gains within only 10 optimization attempts, but with higher variance across attempts and substantial token consumption.

\begin{figure}[h]
    \centering
    \includegraphics[width=\linewidth]{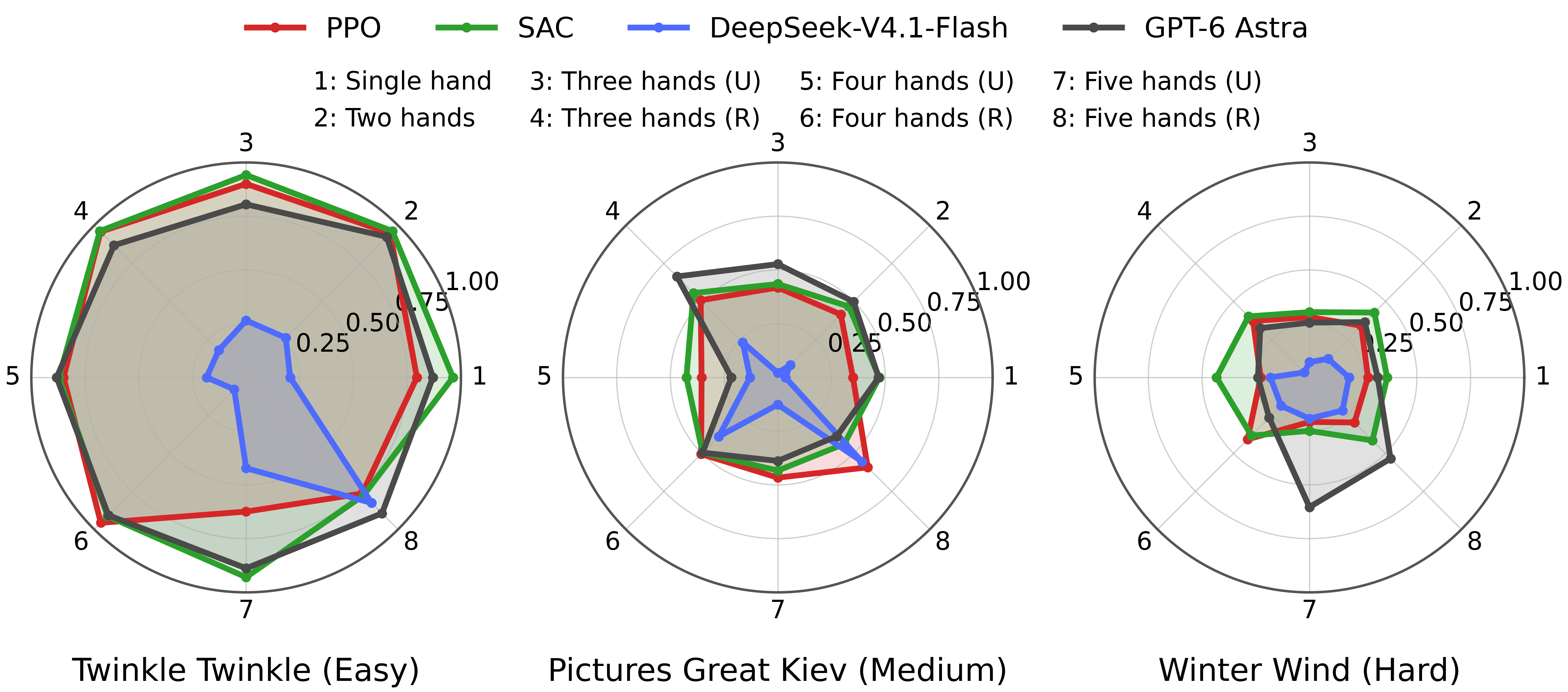}
    \caption{F1-score comparison of PPO, SAC, DeepSeek-V4.1-Flash, and GPT-6 Astra across eight hand settings on three piano pieces. The details of F1 metric can be found in Appendix~\ref{app:evaluation_metric}. }
    \label{fig:general-llm-f1-radar}
\end{figure}

\begin{tcolorbox}[
width=\linewidth,
colback=white!97!black,
colframe=black!80,
boxrule=1.0pt,
arc=2mm,
outer arc=2mm,
boxsep=4pt,
left=4pt,
right=4pt,
top=3pt,
bottom=3pt
]
\itshape
\textbf{Findings:}
\begin{itemize}[leftmargin=*]
\item \textbf{Song duration alone does not explain task difficulty.} Note density is more informative, while difficulty also depends on fingering complexity, hand coordination, and wide leaps.

\item \textbf{More hands do not guarantee better F1 score}, suggesting that additional control freedom can increase exploration and coordination difficulty.

\item \textbf{Both RL and LLM-based agents struggle on \ourbenchmark}, but exhibit distinct optimization behaviors. RL improves more steadily, whereas LLM-based agents can achieve rapid F1 gains within a few attempts at substantial token cost.

\end{itemize}
\end{tcolorbox}

\subsection{Evaluation of Robust RL Baselines}

\label{sec:robust_rl_results}
\begin{figure}[H]
    \centering
    \includegraphics[width=0.8\linewidth]{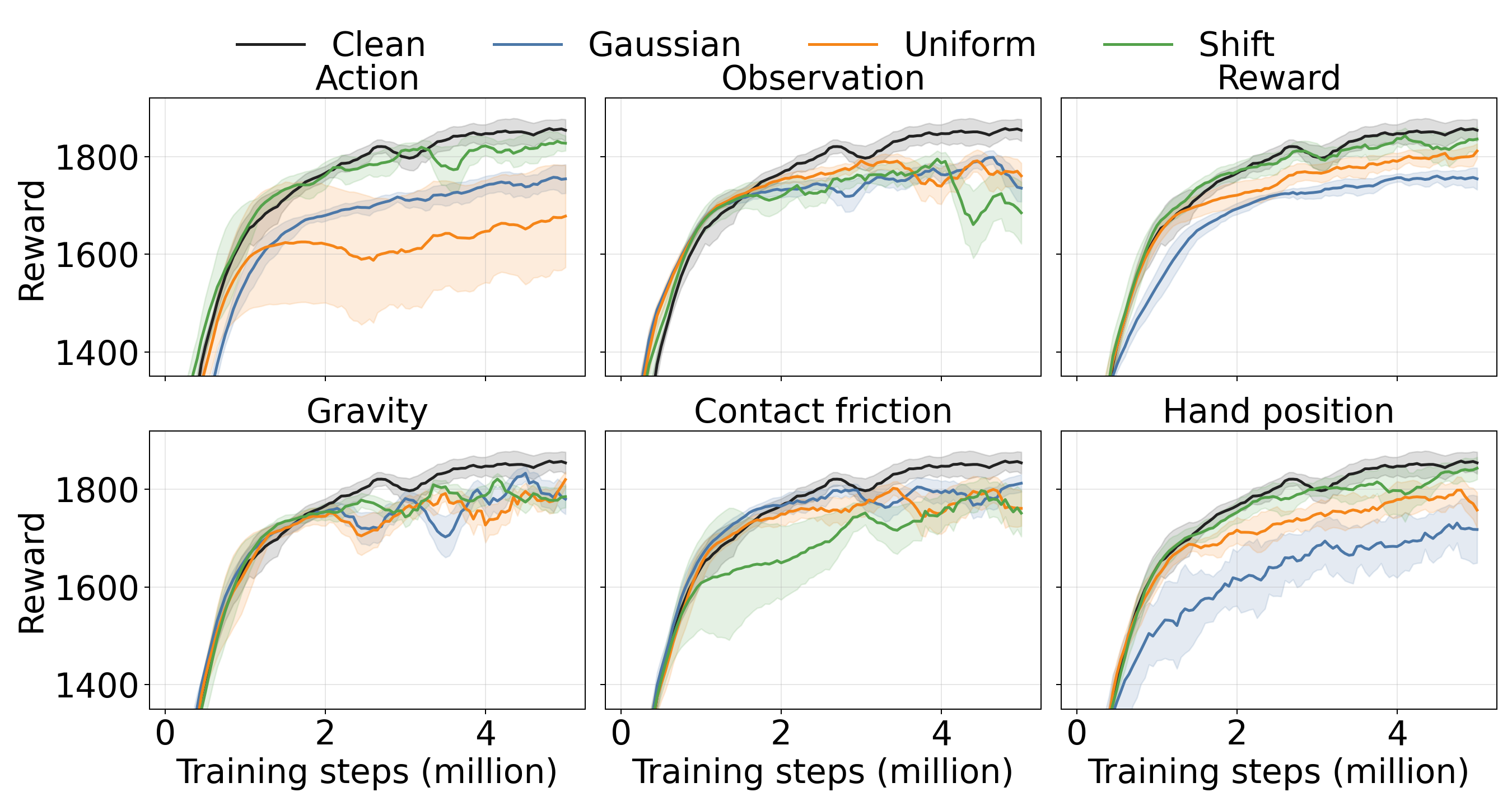}
    \caption{Periodic evaluation reward of PPO across six
      robustness channels under Gaussian, uniform, and shift
      perturbations, with the clean environment shown as a
      reference.}
    \label{fig:ppo_robust_channels_reward}
\end{figure}

In this section, we evaluate robust RL under different perturbation channels, distributions, and magnitudes on two tasks: two-hand \textit{Clair de Lune} and three-hand \textit{Für Elise}. We begin by evaluating PPO under perturbations to action, observation, reward, gravity, contact-friction, and hand-pose, considering Gaussian, uniform, and shift distributions to characterize sensitivity across different disturbance (Figure~\ref{fig:ppo_robust_channels_reward}). Among these perturbations, action and hand-pose disturbances produce the largest performance degradation, while contact-friction and gravity are generally less disruptive. 

We then compare PPO, SAC, A2PSAC~\citep{liu2024robustdeepreinforcementlearning}, SCPO~\citep{kuang2021learningrobustpolicydisturbance}, and EPPO~\citep{akgül2025overcomingnonstationarydynamicsevidential} on two-hand \textit{Clair de Lune} under Gaussian action and reward perturbations in two separate experiments (Figure~\ref{fig:cross_algorithm_gaussian}). A2PSAC and SAC maintain higher performance, whereas EPPO performs weakest overall, revealing substantial differences in robustness across algorithms. Finally, we move beyond isolated perturbations and evaluate composable disturbances on three-hand \textit{Für Elise}, progressively combining Observation (O), Reward (R), Action (A), Gravity (G), and Hand-Pose (HP) perturbations to assess robustness under increasingly complex disturbance combination (Figure~\ref{fig:robust_rl_composable_perturbations}). Detailed analysis of algorithms and disturbance compositions are provided in Appendix~\ref{app:detailed_robustrl}.

\begin{tcolorbox}[
width=\linewidth,
colback=white!97!black,
colframe=black!80,
boxrule=1.0pt,
arc=2mm,
outer arc=2mm,
boxsep=4pt,
left=4pt,
right=4pt,
top=3pt,
bottom=3pt
]
\itshape
\textbf{Finding:}
\textbf{Action and hand-pose perturbations are most disruptive}, highlighting actuation and initialization as key robustness challenges.

\end{tcolorbox}

\begin{figure}[h]
\centering
\begin{subfigure}[t]{0.9\linewidth}
  \centering
  \includegraphics[width=\linewidth]{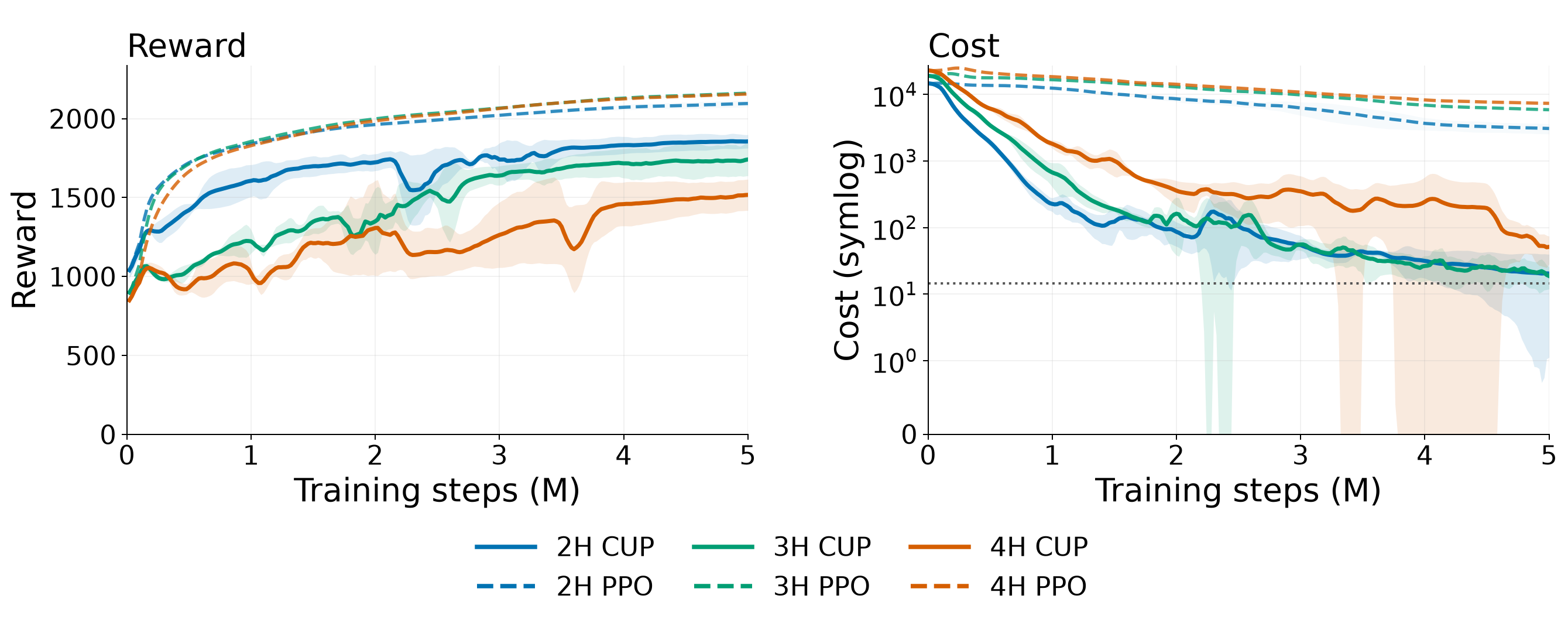}
  \caption{Hand count sensitivity of CUP.}
  \label{fig:main safety_PicturesGreatKiev}
\end{subfigure}
\hfill
\begin{subfigure}[t]{0.9\linewidth}
  \centering
  \includegraphics[width=\linewidth]{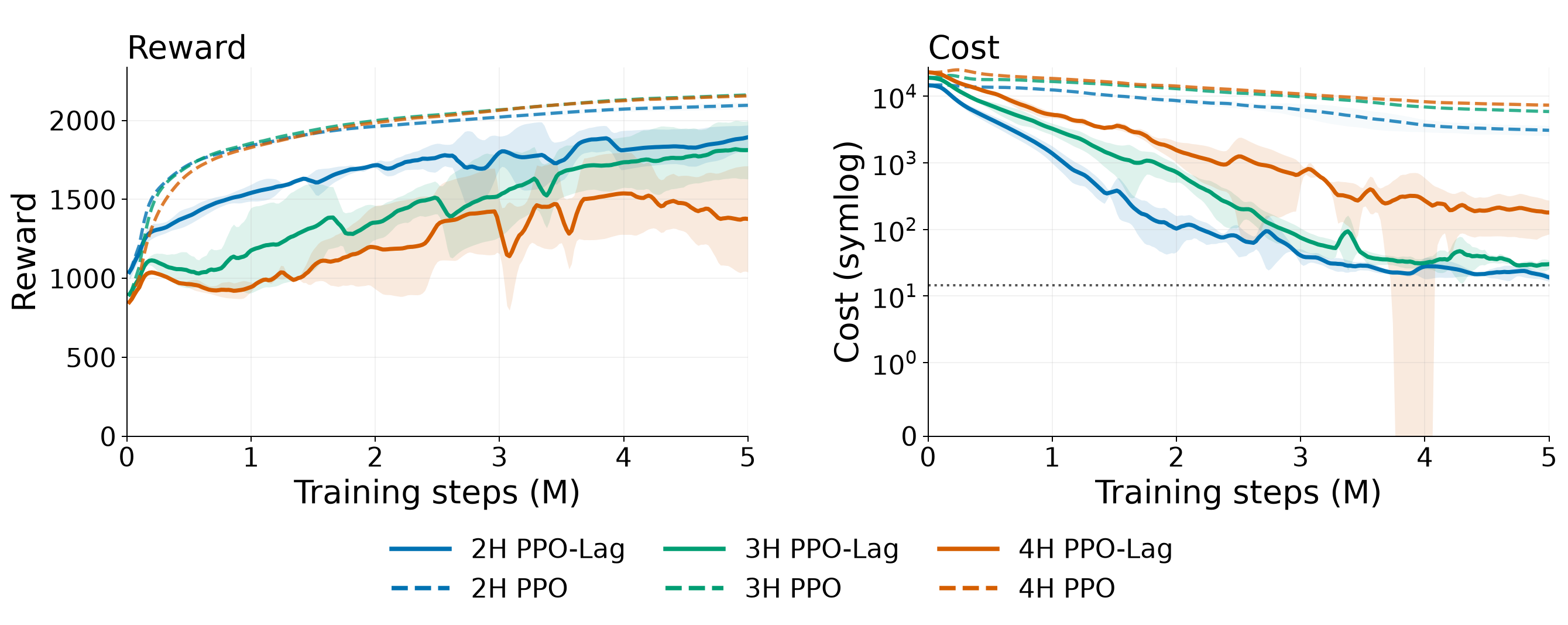}
  \caption{Hand count sensitivity of PPO-Lag.}
  \label{fig:main safety_PolonaiseOp40No1}
\end{subfigure}
\caption{
Effect of hand count on safety RL:
training reward and safety cost for 2 to 4 hands.
Solid lines denote CUP / PPO-Lag; dashed lines denote unconstrained PPO.
The dotted line marks the cost budget. In this task, four hands policies achieve lower
rewards and show greater difficulty reducing safety costs
than two or three hands policies.}
\label{fig:main safety}
\end{figure}

\subsection{Evaluation of Safe RL Baselines}
\label{sec:safe_rl_results}

We compare four safe RL algorithms, PPO-Lag, OnCRPO~\citep{xu2021crpo},
CPO~\citep{achiam2017constrained}, and CUP~\citep{yang2022cup} on \ourbenchmark tasks with two to four hands, using PPO as a reference.
All methods use the OT reward and are trained for 5M
environment steps with three random seeds, as shown in Figure \ref{fig:main safety} and Appendix \ref{detailed experiments safety}.
We examine reward and safety cost jointly to assess
whether agents learn to reduce constraint violations
while maintaining piano playing performance.
Figure~\ref{fig:main safety} shows hand count sensitivity results for CUP and PPO-Lag;
results for OnCRPO and CPO are provided in
Appendix~\ref{additional hand count results}. The key conclusion is summarized in the finding.

We further study sensitivity to the cost budget, as shown in Appendix \ref{cost list section}.
The budget study varies the cost limit while keeping
the task fixed, examining how each algorithm balances
reward and safety.
The hand count study compares two, three, and four hands
configurations with the same piece, safety semantic,
and cost definition.
Additional results and experimental settings are provided
in Appendix ~\ref{detailed experiments safety} and ~\ref{detailed safety parameters}.

\begin{tcolorbox}[
width=\linewidth,
colback=white!97!black,
colframe=black!80,
boxrule=1.0pt,
arc=2mm,
outer arc=2mm,
boxsep=4pt,
left=4pt,
right=4pt,
top=3pt,
bottom=3pt
]
\itshape
\label{safety finding}
\textbf{Finding:}
\textbf{Increasing the number of hands can make safe RL more difficult}. In the evaluated task, policies controlling four hands struggle more to reduce safety costs and achieve lower rewards than those controlling two or three hands.

\end{tcolorbox}

\begin{figure}[h]
    \centering
    \includegraphics[
        width=0.8\textwidth,
        trim=0 10 0 10,
        clip
    ]{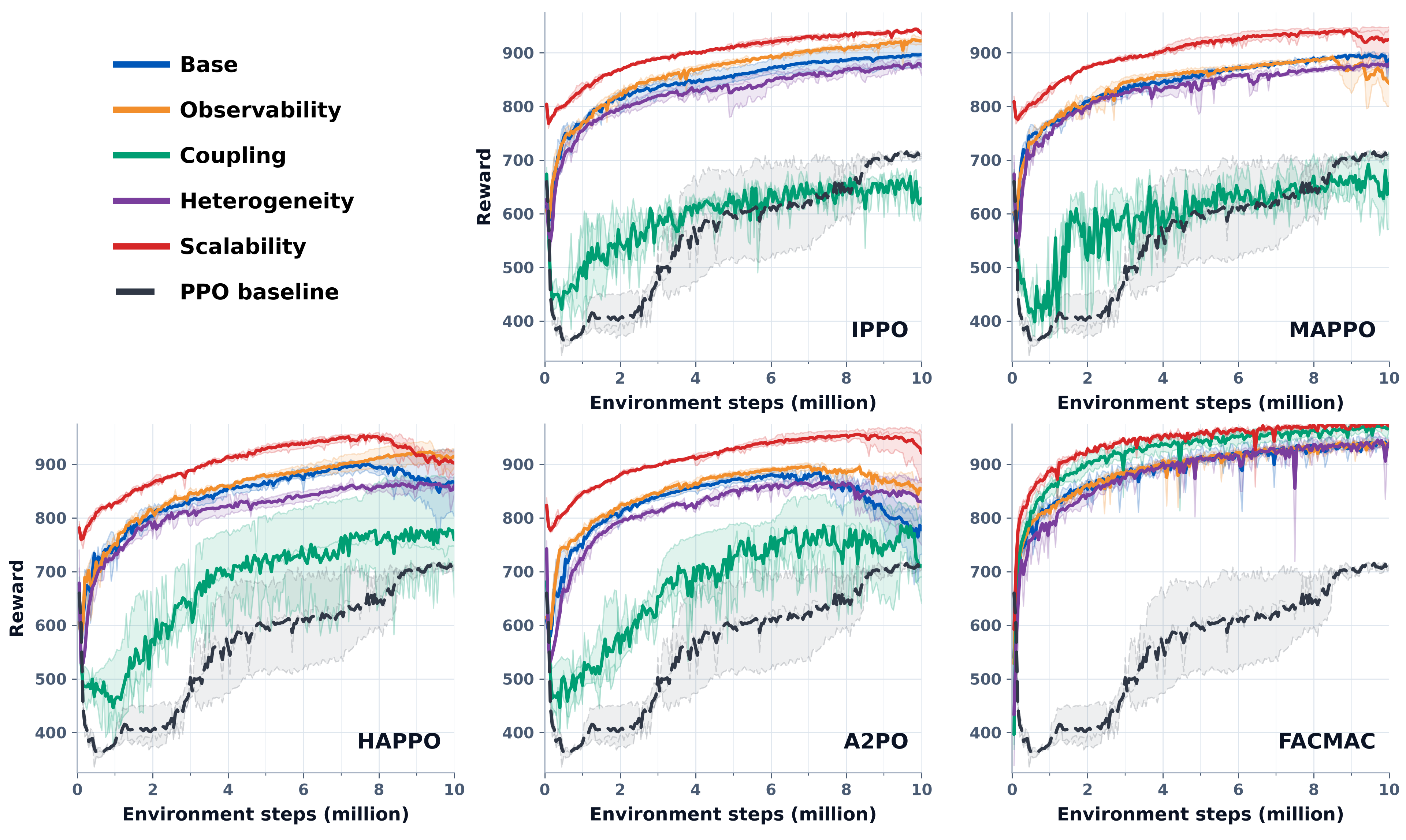}
    \caption{Task-wise training performance of representative MARL algorithms across the five settings. Compared with Base, Scalability and Observability variant generally improve performance, while Heterogeneity and especially Coupling variant make policy learning more difficult. Among the variants, Coupling most clearly differentiates the algorithms.}
    \label{fig:marl_algorithm_comparison}
\end{figure}

\subsection{Evaluation of Multi-Agent RL Baselines}
\label{sec:marl_results}
We evaluate representative MARL algorithms under a base task and four SCHO variants. The tasks are build on the same four-hand piece, \textit{WinterWind}. Specifically, the \textbf{Base task} uses two agents with a balanced two-hand assignment, disjoint keyboard action ranges, and localized partial observations; the \textbf{Observability variant} keeps the same balanced assignment and action ranges as Base, but provides broader local observations; the \textbf{Coupling variant} uses two balanced agents with overlapping keyboard action ranges while retaining the Base observation ranges; the \textbf{Heterogeneity variant} assigns one hand to one agent and three hands to the other, creating asymmetric control loads and action dimensions; and the \textbf{Scalability variant} uses four agents to control the same four hands, with one hand assigned to each agent. All settings share the same physical hands, MIDI objective, team reward, and training budget. For algorithms, we compare IPPO\citep{dewitt2020ippo}, MAPPO\citep{yu2022mappo}, HAPPO\citep{kuba2022happo}, A2PO\citep{wang2023order}, and FACMAC\citep{peng2021facmac}, and include centralized PPO\citep{schulman2017proximalpolicyoptimizationalgorithms} as a base reference with a single agent accessing to the full observation and action space. Figure~\ref{fig:marl_algorithm_comparison} shows the task-wise evaluation return for each algorithm. Evaluation uses the reward in Equation~\ref{eq:ot reward} for benchmark consistency, while during training we add an additional shared per-step penalty for inter-agent hand contact. Detailed task setup and experiment analysis are provided in Appendix~\ref{app:marl_exp_result}.

\begingroup
\renewcommand{\thefootnote}{*}

\begin{tcolorbox}[
width=\linewidth,
colback=white!97!black,
colframe=black!80,
boxrule=1.0pt,
arc=2mm,
outer arc=2mm,
boxsep=4pt,
left=4pt,
right=4pt,
top=3pt,
bottom=3pt
]
\itshape
\textbf{Finding: More agents can be beneficial for learning dexterous piano playing\footnotemark{}.}
With the same four-hand setting, performance improves from centralized PPO with one agent, to the two-agent Base setting, and further to the four-agent Scalability setting. This suggests that distributing the task across more agents can lead to better learning.
\end{tcolorbox}

\footnotetext{Here, ``more agents'' means more controllers, not increasing the number of hands or the total action space. Instead, the term ``more hands'' used earlier refers to increasing the number of hands and the corresponding action space of the task.}

\endgroup

\section{Conclusion}
\label{sec:conclusion}
We introduced \ourbenchmark, a unified benchmark for dexterous piano playing that supports one to five hands, up to 111 continuous action dimensions, 912 task settings, and 36 baseline algorithms across standard, robust, safe, and multi-agent RL. By integrating scalable control, configurable perturbations, explicit safety constraints, decentralized cooperation, and LLM-based agents within a shared task family, \ourbenchmark enables systematic evaluation beyond task performance. Our experiments show that challenging piano-playing tasks remain difficult for current RL methods and frontier LLM-based agents, even under standard conditions. Together with its modular architecture and public release, \ourbenchmark has the potential to become a versatile testbed for evaluating and developing diverse RL
algorithms in dexterous control.

\bibliography{references}
\bibliographystyle{references}

\section*{Reproducibility Statement}

We provide the information required to reproduce \ourbenchmark and the reported experiments throughout the main text and appendices. The benchmark task formulation, observation and action spaces, reward design, and evaluation metrics are described in Section~\ref{sec:benchmark}, Section~\ref{sec:experiments} and  Appendix~\ref{app:additional_experiments}. Experimental protocols, baseline implementations, and hyperparameter settings are documented in Appendix~\ref{app:experimental_settings}. The accompanying code includes the benchmark environments, configuration files, and scripts for reproducing the reported experiments.

\section*{Limitations and Broader Impact}

\ourbenchmark is limited to dexterous piano playing in simulation, and its findings may not directly generalize to other manipulation tasks or real-world robotic systems. Results may also depend on the simulated morphology, contact dynamics, and task-specific perturbations and constraints. Nevertheless, \ourbenchmark provides a unified platform for studying high-dimensional dexterous control, robustness, safety, and multi-agent coordination under consistent tasks and evaluation protocols.

\appendix

\section{Preliminaries}
\label{app:background_problem}

\ourbenchmark is built upon a common dexterous piano-playing formulation, in which robotic hands interact with a simulated piano to follow a MIDI-conditioned target sequence. This section introduces the shared state, observation, action, and reward definitions, and then formalizes standard RL as the base setting from which the robust, safe, and multi-agent variants are derived.

\subsection{Standard Reinforcement Learning}
\label{subsec:standard_rl}

Given a MIDI score $m$ with horizon $H_m$, the standard piano-playing task
is formulated as a finite-horizon Markov Decision Process (MDP):
\begin{equation}
    \mathcal{M}_m =
    \left(
        \mathcal{S}, \mathcal{A}, P_m, r_m, \rho_0, \gamma, H_m
    \right),
\end{equation}
where $\mathcal{S}$ and $\mathcal{A}$ denote the state and action spaces,
$\rho_0$ is the initial-state distribution, $P_m$ is the transition kernel,
and $r_m$ is the reward function.
Let $\mathcal{H}=\{1,\ldots,N_h\}$ denote the set of robotic hands.
The policy receives an observation $o_t := h(s_t)\in\mathcal{O}$ consisting of
hand proprioception, piano-key states, and score-conditioned goals, while
the action $a_t\in\mathcal{A}$ specifies the target joint positions of all
hands and the sustain-pedal command:
\begin{equation}
    \begin{aligned}
    o_t &=
    \left[
        \{q_t^{h}, x_t^{h}\}_{h\in\mathcal{H}},
        k_t,
        g_{t:t+L-1}
    \right], \\
    a_t &=
    \left[
        \{a_t^{h}\}_{h\in\mathcal{H}},
        a_t^{\mathrm{pedal}}
    \right].
    \end{aligned}
\end{equation}
where $q_t^{h}$ and $x_t^{h}$ denote the joint configuration and Cartesian
forearm pose of hand $h$, respectively, $k_t$ denotes the piano-key joint
states, and $g$ denote the key-activation targets over a lookahead horizon $L$. Under this formulation, the policy is optimized by
\begin{equation}
    \pi^\star =
    \arg\max_{\pi} J_R(\pi),
    \qquad
    J_R(\pi) =
    \mathbb{E}_{\tau\sim\pi}
    \left[
        \sum_{t=0}^{H_m-1}\gamma^t r_t
    \right].
    \label{eq:standard_rl}
\end{equation}

Here, $r_t$ measures the agent's piano-playing performance at time $t$ in terms of key accuracy, sustain control, and fingertip-to-key matching.

\subsection{Robust Reinforcement Learning with Disruptions}
\label{subsec:robust_rl}

Starting from the standard MDP, we formulate robustness using the modular
disrupted-interaction framework of Robust-Gymnasium
\citep{gu2025robustgymnasium}. Let:
\begin{equation}
    \mathcal{M}_m^0 =
    \left(
        \mathcal{S},\mathcal{A},
        P_{m,\phi_0},r_{m,\phi_0},
        \rho_{\phi_0},\gamma,H_m
    \right)
\end{equation}
denote the nominal piano-playing task, where $\phi_0$ contains the nominal
environment parameters.

A disrupted interaction is described by observation, action, reward, and
environment operators $D_o$, $D_a$, $D_r$, and $D_e$:
\begin{equation}
\begin{aligned}
    \tilde{o}_t
        &= D_o(o_t;\xi_t^o),
    & a_t
        &= \sim \pi(\cdot\mid\tilde{o}_t),
    & \tilde{a}_t
        &= D_a(a_t;\xi_t^a),\\
    \phi_t
        &= D_e(\phi_0;\xi_t^e),
    & s_{t+1}
        &\sim P_{m,\phi_t}
        (\cdot\mid s_t,\tilde{a}_t),
    & \tilde{r}_t
        &= D_r\!\left(
            r_{m,\phi_t}(s_t,\tilde{a}_t,s_{t+1});
            \xi_t^r
        \right).
\end{aligned}
\label{eq:disrupted_interaction}
\end{equation}

The resulting objective is:

\begin{equation}
    \pi^\star =
    \arg\max_{\pi} J_R(\pi),
    \qquad
    J_R(\pi) =
    \mathbb{E}_{\tau,\xi}
    \left[
        \sum_{t=0}^{H_m-1}
        \gamma^t\tilde{r}_t
    \right].
    \label{eq:robust_objective}
\end{equation}

Here, $D_o$, $D_a$, and $D_r$ corrupt signals during interaction, whereas
$D_e$ may modify the transition dynamics or initial-state distribution.
Setting all disruption operators to the identity recovers the standard
piano-playing MDP.

\subsection{Safe Reinforcement Learning}
\label{subsec:safe_rl}

Safe piano playing is formulated as a constrained Markov decision process
(CMDP) \citep{gu2024review}:
\begin{equation}
    \mathcal{M}_m^C =
    \left(
        \mathcal{S},\mathcal{A},P_m,r_m,c,
        \rho_0,\gamma,H_m,d
    \right),
\end{equation}
where $c_t=c(s_t,a_t,s_{t+1})\geq 0$ is the safety cost and $d$ is the
episodic cost limit. If multiple cost components are enabled, they are
aggregated as:
\begin{equation}
    c_t=\sum_{k=1}^{K}\alpha_k c_{k,t}.
\end{equation}

The Safe-RL objective is:
\begin{equation}
\begin{aligned}
    \max_{\pi}\quad&
    J_R(\pi)
    =
    \mathbb{E}_{\tau\sim\pi}
    \left[
        \sum_{t=0}^{H_m-1}\gamma^t r_t
    \right],\\
    \mathrm{s.t.}\quad&
    J_C(\pi)
    =
    \mathbb{E}_{\tau\sim\pi}
    \left[
        \sum_{t=0}^{H_m-1}c_t
    \right]
    \leq d.
\end{aligned}
\label{eq:safe_rl}
\end{equation}

\ourbenchmark considers safety costs related to excessive control or joint
motion, inter-hand contact, actuator power, and joint-specific risk.
Reward and cost are maintained as separate signals, so a policy must
improve musical performance while satisfying the safety constraint.
Benchmark feasibility is determined using the undiscounted episodic cost
in Eq.~\eqref{eq:safe_rl}.

\subsection{Multi-Agent Reinforcement Learning}
\label{subsec:multiagent_rl}

The multi-agent extension distributes the piano-playing task among
multiple agents and is modeled as a cooperative partially observable
Markov game: 
\begin{equation}
    \mathcal{G}_m =
    \left(
        \mathcal{I},\mathcal{S},
        \{\mathcal{O}_i\}_{i\in\mathcal{I}},
        \{\mathcal{A}_i\}_{i\in\mathcal{I}},
        P_m,r_m,\rho_0,\gamma,H_m
    \right),
\end{equation}
where $\mathcal{I}$ is the set of agents. Each agent selects an action from
its local observation, and the joint policy factorizes as: 
\begin{equation}
    \boldsymbol{\pi}
    (\boldsymbol{a}_t\mid\boldsymbol{o}_t)
    =
    \prod_{i\in\mathcal{I}}
    \pi_i(a_t^i\mid o_t^i).
\end{equation}

All agents share the same team reward and optimize:
\begin{equation}
    \boldsymbol{\pi}^{\star}
    =
    \arg\max_{\boldsymbol{\pi}}
    \mathbb{E}_{\tau\sim\boldsymbol{\pi}}
    \left[
        \sum_{t=0}^{H_m-1}\gamma^t r_t
    \right].
    \label{eq:multiagent_rl}
\end{equation}

A constrained Markov game could additionally introduce agent-specific or
team-level safety costs and budgets. However, the current \ourbenchmark
benchmark evaluates safety and multi-agent learning as separate task
families and does not yet combine them into a safe multi-agent setting.

\section{Related Work}
\label{app:related_work}

\subsection{Robotic Piano-Playing and Benchmarks}
\label{app:piano_benchmarks}

The closest foundations for \ourbenchmark are simulated, score-conditioned piano-playing tasks. Early work formulated the problem as reinforcement learning for dexterous hands~\citep{xu2022learningplaypianodexterous}, while RoboPianist established a reproducible MuJoCo benchmark with two Shadow Hands, 150 pieces, and note-level metrics~\citep{zakka2023robopianist}. These works primarily evaluate nominal bimanual performance, leaving control beyond two hands and evaluation under perturbations, safety constraints, or decentralized control outside their scope.

Subsequent studies have extended individual aspects of piano playing: RP1M and PianoMime scale demonstration data and cross-song generalization~\citep{pmlr-v270-zhao25d,pmlr-v270-qian25a}; real-world systems study dynamics calibration, domain randomization, and residual adaptation~\citep{zeulner2026learningplaypianoreal,xie2026handelbotrealworldpianoplaying}; and CANTABILE enriches musical evaluation with strike-velocity accuracy~\citep{kim2026cantabilelearningexpressivedynamics}. These advances remain separated across data scale, physical deployment, and musical evaluation. \ourbenchmark instead brings multi-hand scaling, controlled perturbations, explicit safety costs, and decentralized cooperation into a shared piano-playing benchmark.

\subsection{Standard Reinforcement Learning Benchmarks}
\label{subsec:related_general_rl}
Standard RL benchmarks cover a wide range of evaluation settings. General-purpose and data-oriented benchmarks standardize tasks, datasets, and generalization protocols~\citep{bellemare2013arcade,cobbe2020leveragingproceduralgenerationbenchmark,fu2021d4rldatasetsdeepdatadriven,gulcehre2021rlunpluggedsuitebenchmarks,qin2021neorlnearrealworldbenchmark}. Continuous-control benchmarks provide standardized physics-based environments and interfaces~\citep{brockman2016openai,tassa2018deepmind,sferrazza2024humanoidbenchsimulatedhumanoidbenchmark}. Robotic manipulation and specialized benchmarks further study multi-task learning, generalization, safety, and robustness~\citep{yu2020meta,zhu2020robosuite,gu2023maniskill2,ji2023neurips-safety,pumacay2024colosseumbenchmarkevaluatinggeneralization}.\ourbenchmark complements these efforts with a unified dexterous piano-playing task family spanning standard, robust, safe, and multi-agent RL.

\subsection{Robust Reinforcement Learning Benchmarks}
\label{app:robust_rl_related_work}
Robust RL studies uncertainty entering different components of the agent--environment interaction, including observations, actions, transition dynamics, rewards or costs, timing, and visual or semantic task factors~\citep{huang2017adversarialattacksneuralnetwork,pmlr-v97-tessler19a,pmlr-v70-pinto17a,Sim2Real2018,bouteiller2021iclr-reinforcement,tobin2017domain}. Such uncertainties may also evolve over time or occur jointly~\citep{lecarpentier2019nonstationary,mandlekar2017adversarial}.

Among benchmarks explicitly targeting robust RL, RRLS~\citep{zouitine2024rrlsrobustreinforcement} standardizes evaluation on six MuJoCo tasks with uncertain physical parameters and external forces, while Robust-Gymnasium~\citep{gu2025robustgymnasium} provides modular disruptions across a broad task collection. Robustness tests also appear in broader RL benchmarks and algorithm papers~\citep{duan2016benchmarking,pmlr-v70-pinto17a,pmlr-v97-tessler19a,NEURIPS2020_f0eb6568,kuang2021learningrobustpolicydisturbance,NEURIPS2023_d0c38418}, but they are typically one-off evaluations with task- and method-specific protocols, limiting direct comparison. In contrast, \ourbenchmark evaluates robustness in piano playing, where high-dimensional continuous control and multi-hand coordination provide a more demanding testbed for revealing differences in algorithmic robustness.

\subsection{Safe Reinforcement Learning Benchmarks}
\label{subsec:related_work_safety}

Safe RL is commonly formulated as a constrained Markov decision process, where policies optimize task return subject to safety-cost constraints~\citep{garcia2015comprehensive, berkenkamp2017safe,yu2025safe,konighofer2025shields}. Representative methods include CPO~\citep{achiam2017constrained} and primal-dual~\citep{paternain2022safe, bai2022achieving, xu2025global,pengtowards} or Lagrangian approaches. Benchmark suites such as Safety Gym~\citep{ray2019benchmarking}, Safety-Gymnasium~\citep{ji2023neurips-safety}, GUARD~\citep{zhao2023guard}, and OmniSafe~\citep{ji2024omnisafe} provide standardized environments and evaluation tools for studying safety in navigation, locomotion, and manipulation, while safe multi-agent RL extends these ideas to constrained Markov games~\citep{gu2023safe}. Compared with prior Safe-RL benchmarks, \ourbenchmark focuses on long-horizon dexterous piano playing and evaluates safety under physically meaningful constraints while preserving the underlying musical objective.

\subsection{Multi-Agent Reinforcement Learning and Benchmarks}
\label{app:marl_related_work}
Multi-agent Reinforcement Learning (MARL) studies coordination among decentralized policies under a shared objective. IPPO~\citep{dewitt2020ippo} provides a simple decentralized baseline, while methods such as MADDPG~\citep{lowe2017maddpg}, COMA~\citep{foerster2018coma}, QMIX~\citep{rashid2018qmix}, MAPPO~\citep{yu2022mappo}, and FACMAC~\citep{peng2021facmac} use Centralized information during Training while retaining Decentralized Execution (CTDE). Other methods target specific challenges, including Mean Field MARL~\citep{yang2018meanfield} for large agent populations, HAPPO~\citep{kuba2022happo} for heterogeneous agents, and MAT~\citep{wen2022mat} for sequence-based multi-agent coordination. Together, these methods highlight scalability, coupling, heterogeneity, and observability as recurring challenges in MARL.

Existing MARL benchmarks emphasize different subsets of these challenges. MAgent~\citep{zheng2018magent} and Neural MMO~\citep{suarez2019neuralmmo} focus on large agent populations; SMAC~\citep{samvelyan2019smac}, cooperative MPE~\citep{lowe2017maddpg}, and Google Research Football~\citep{kurach2020grf} emphasize coordinated interaction; Hanabi~\citep{bard2020hanabi} and SMACv2~\citep{ellis2023smacv2} emphasize partial observability; and CH-MARL~\citep{sharma2022chmarl} introduces heterogeneous agents. Multi-Agent MuJoCo~\citep{dewitt2020mujoco} and Bi-DexHands~\citep{chen2023bidexhands} further extend MARL evaluation to continuous robotic control and dexterous manipulation. These benchmarks typically cover only a subset of scalability, coupling, heterogeneity, and observability, and some entangle multiple factors together. \ourbenchmark instead makes all four SCHO dimensions separately configurable within the same dexterous-control environment, enabling controlled comparison under a shared embodiment and musical objective.

\section{Additional Experiments and Analysis}
\label{app:additional_experiments}

\subsection{Standard RL}
\label{sec:additional_standard_rl_results}
\subsubsection{Sweep study.} We examine the relationship between song difficulty and task performance through a sweep study across representative piano pieces (Table~\ref{tab:song_difficulty_f1}). Here, NPS denotes note density in notes per second, while steps is computed at a $20\,\mathrm{Hz}$ control frequency. Table~\ref{tab:song_difficulty_f1} shows that song duration alone does not explain task difficulty. Short but note-dense pieces such as \textit{Winter Wind} can be substantially 

\begin{table}[H]
\caption{Task statistics and performance across piano pieces.}
\label{tab:song_difficulty_f1}
\centering
\small
\setlength{\tabcolsep}{4pt}
\renewcommand{\arraystretch}{1.08}
\begin{tabularx}{\linewidth}{@{}Xrrr@{}}
\toprule
Song & Steps & NPS & F1 Score \\
\midrule
Gymnopédie No. 1                       & 657  & 1.77  & 0.88 \\
Twinkle Twinkle (Rousseau)             & 451  & 1.51  & 0.86 \\
Clair de Lune                           & 597  & 2.65  & 0.63 \\
French Suite No. 1, Allemande          & 517  & 10.89 & 0.59 \\
French Suite No. 5, Sarabande          & 962  & 3.56  & 0.59 \\
Piano Sonata No. 23, 2nd mov.          & 313  & 12.80 & 0.49 \\
Pictures at an Exhibition: Great Kiev  & 720  & 8.39  & 0.45 \\
Kreisleriana Op. 16 No. 8              & 591  & 9.78  & 0.44 \\
Piano Sonata No. 2, 1st mov.           & 361  & 6.94  & 0.44 \\
Piano Sonata K279 in C major, 1st mov. & 579  & 9.54  & 0.44 \\
Für Elise                               & 399  & 10.44 & 0.42 \\
French Suite No. 5, Gavotte            & 962  & 6.80  & 0.40 \\
Piano Sonata D845, 1st mov.             & 1000 & 9.04  & 0.34 \\
Winter Wind                             & 314  & 16.64 & 0.28 \\
Bagatelle Op. 3 No. 4                  & 550  & 9.24  & 0.26 \\
Partita No. 2, 6th mov.                 & 425  & 11.91 & 0.25 \\
\bottomrule
\end{tabularx}
\vspace{-2mm}
\begin{flushleft}
\footnotesize
NPS denotes note density in notes per second. Steps are computed at a
$20\,\mathrm{Hz}$ control frequency.
\end{flushleft}
\end{table}

\begin{figure}[H]
    \centering
    \includegraphics[width=1\linewidth]{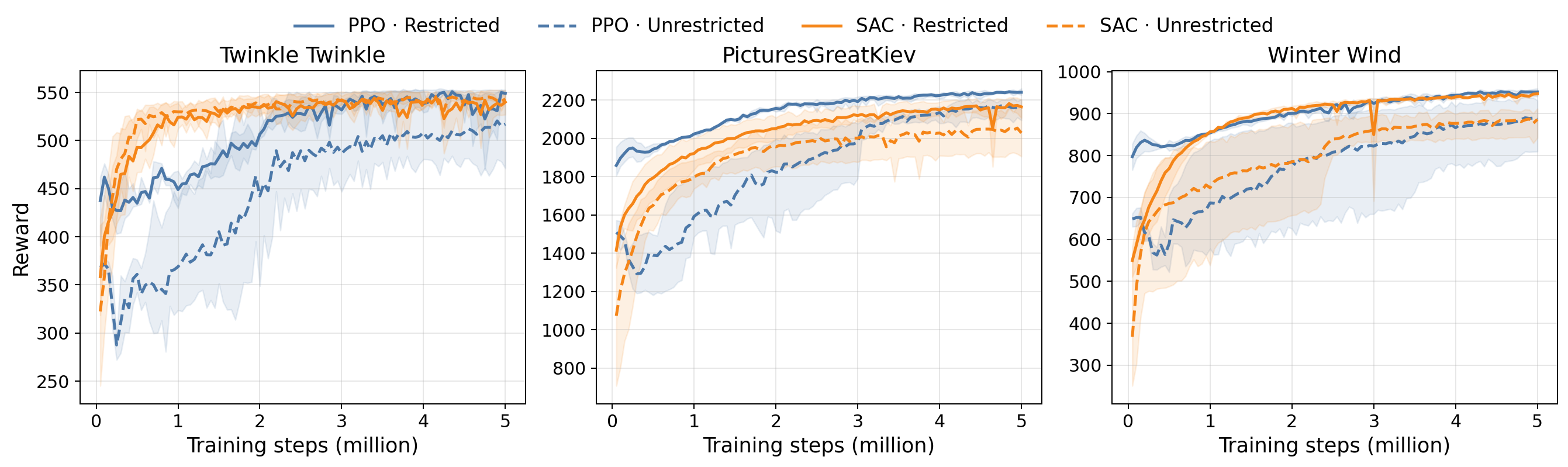}
    \caption{Periodic evaluation reward during training for PPO and SAC under Restricted and Unrestricted configurations across the three songs.}
    \label{fig:general-rl-restricted-learning}
\end{figure}

harder than longer pieces with lower NPS, suggesting that note density is a more informative indicator than episode length. However, the variation among pieces with similar NPS also indicates that difficulty depends on additional factors such as fingering, spatial transitions, and coordination.

\subsubsection{Restricted vs. Unrestricted Learning}
\label{sec:Restricted_vs_Unrestricted learning}
We compare PPO and SAC under restricted and unrestricted hand configurations across the three songs (Figure~\ref{fig:general-rl-restricted-learning}). Restricting each hand to a predefined workspace tends to lead to faster and more stable learning, despite reducing the reachable space. This suggests that the unrestricted setting introduces additional exploration and coordination complexity, while workspace constraints can simplify policy learning by reducing unnecessary motion and inter-hand interference.

\subsection{Robust RL}
\label{app:detailed_robustrl}
\subsubsection{Cross-Algorithm Comparison.}
\begin{figure}[H]
\centering
\begin{subfigure}[t]{0.49\linewidth}
  \centering
  \includegraphics[width=\linewidth]{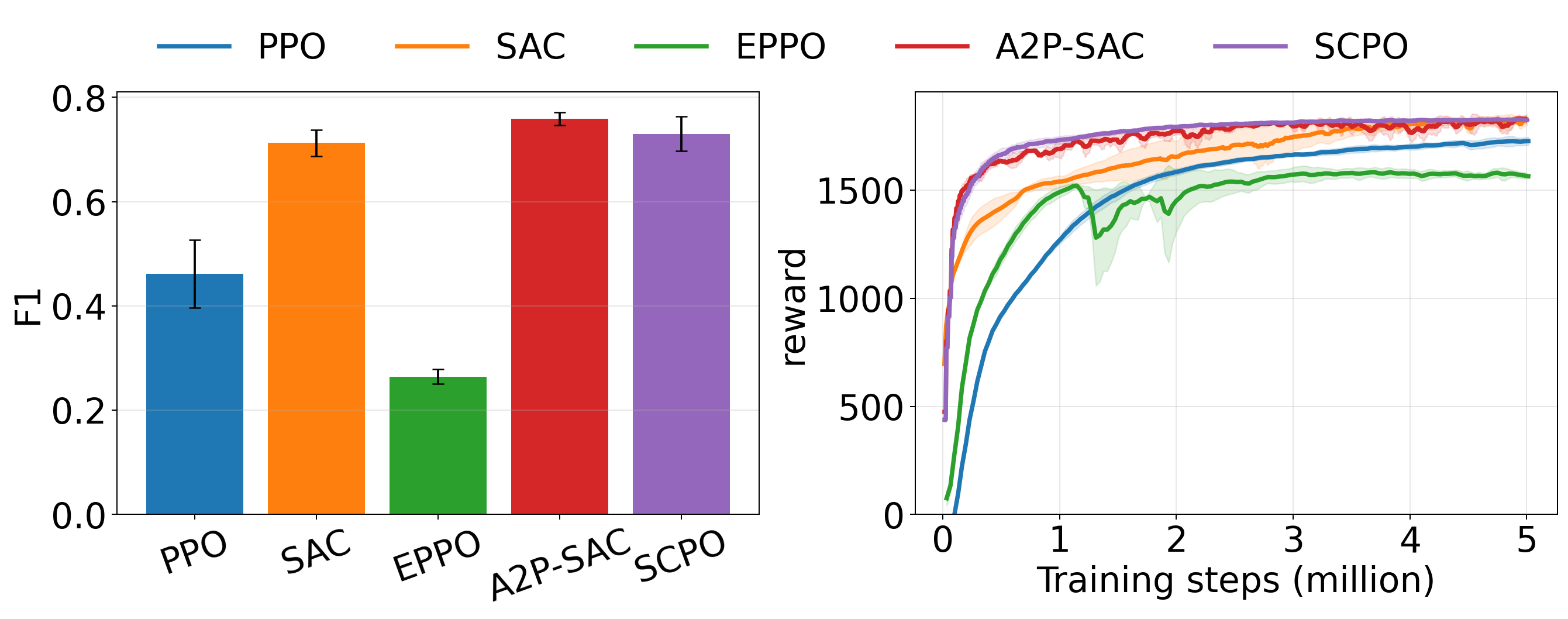}
  \caption{Under Gaussian action perturbations.}
  \label{fig:cross_algorithm_action_gaussian}
\end{subfigure}
\hfill
\begin{subfigure}[t]{0.49\linewidth}
  \centering
  \includegraphics[width=\linewidth]{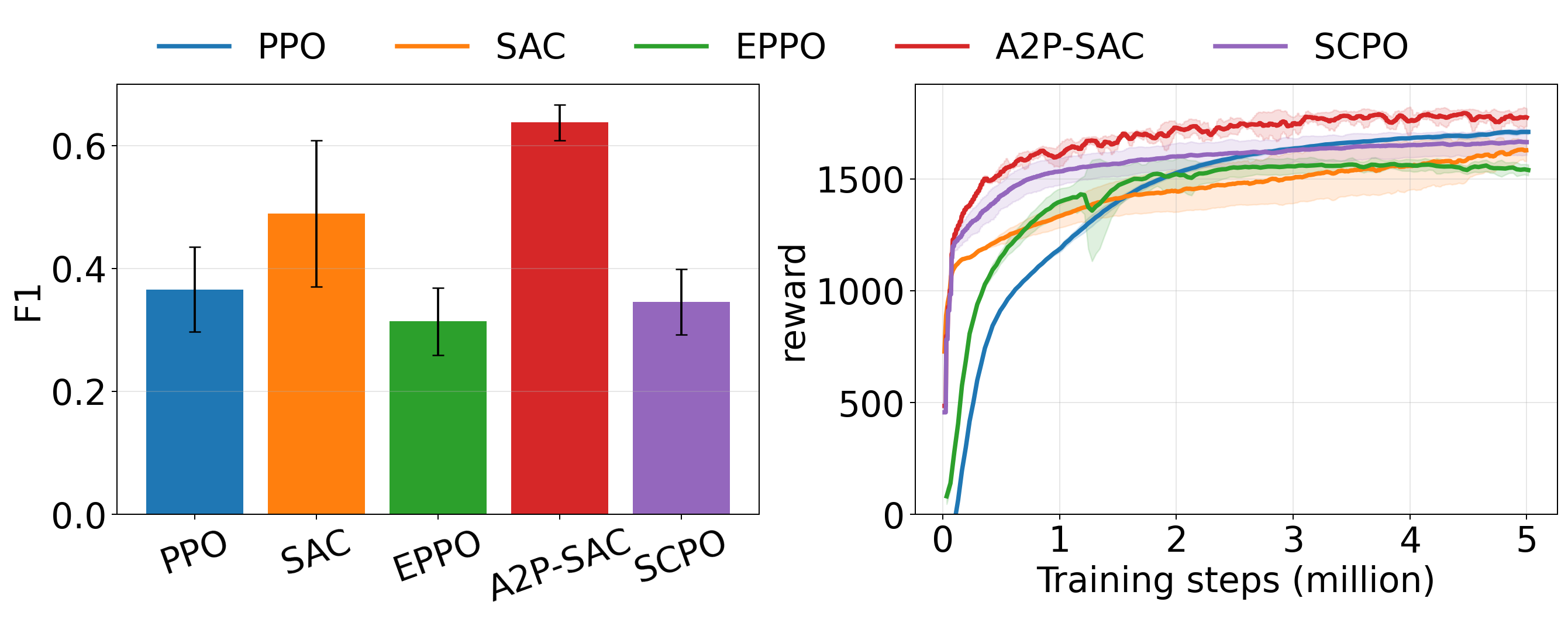}
  \caption{Under Gaussian reward perturbations.}
  \label{fig:cross_algorithm_reward_gaussian}
\end{subfigure}
\caption{
  Cross-algorithm robustness comparison on \textit{Clair de Lune}.
}
\label{fig:cross_algorithm_gaussian}
\end{figure}

We compare five RL methods on two-hand \textit{Clair de Lune} under Gaussian action and reward perturbations. As shown in Figure~\ref{fig:cross_algorithm_gaussian}, the comparison covers Gaussian action perturbations (Figure~\ref{fig:cross_algorithm_action_gaussian}) and Gaussian reward perturbations (Figure~\ref{fig:cross_algorithm_reward_gaussian}). EPPO performs the weakest overall, while A2P-SAC and SAC maintain higher and more stable performance than PPO and SCPO. One possible 

\begin{figure}[H]
    \centering
    \includegraphics[width=\linewidth]{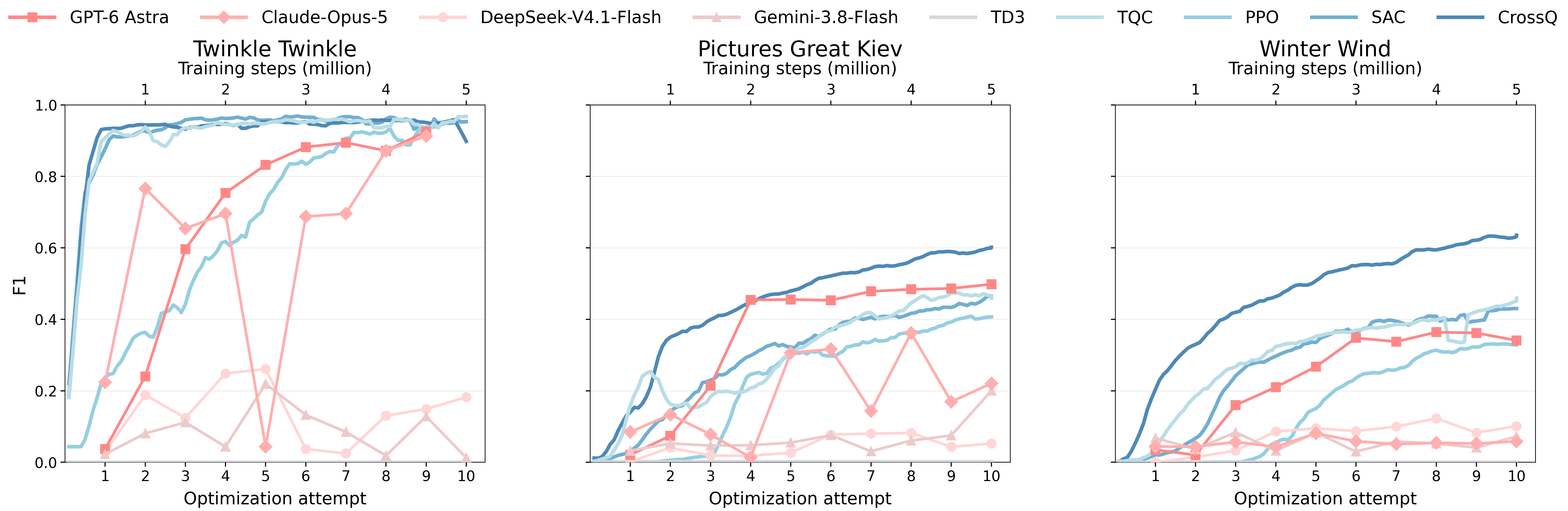}
    \caption{F1 of LLMs over ten keyframe-optimization iterations and RL method over 5M steps for the two-hand setting on three piano pieces.}
    \label{fig:twohand_f1}
\end{figure}

explanation is that A2P-SAC is explicitly designed to improve robustness to action-space perturbations, while SAC also benefits from relatively stable off-policy learning.

\subsubsection{Composable Perturbation Evaluation}
\begin{figure}[H]
      \centering
      \includegraphics[width=\linewidth]{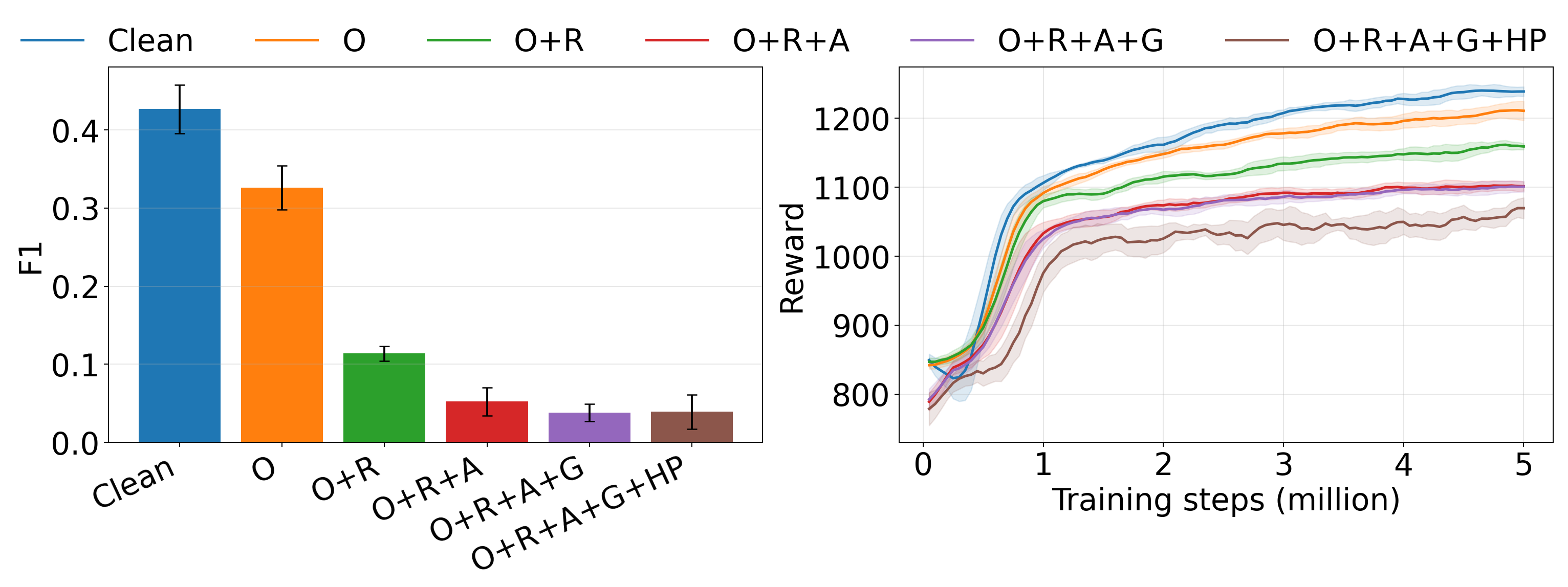}
      \caption{
          Composability of Gaussian perturbations on \textit{Für Elise} with three hands.
      }
      \label{fig:robust_rl_composable_perturbations}
\end{figure}

Moreover, we also demonstrate the ability to leverage different perturbation channels. Figure~\ref{fig:robust_rl_composable_perturbations} shows the performance of PPO under composable Gaussian perturbations from observation (O), observation + reward (O + R), observation + reward + action (O + R + A), observation + reward + action + gravity (O + R + A + G) and observation + reward + action + gravity + hand-pose (O + R + A + G + HP) on \textit{ForElise} with three hands. The results show that the performance of PPO decreases as more perturbation channels are added, indicating that the composability of perturbations can significantly affect the robustness of RL algorithms.

\subsection{Safety RL}
\label{detailed experiments safety}

\subsubsection{Additional Hand-Count Results}
\label{additional hand count results}
\begin{figure}[H]
\centering
\begin{subfigure}[t]{0.49\linewidth}
\centering
\includegraphics[width=\linewidth]{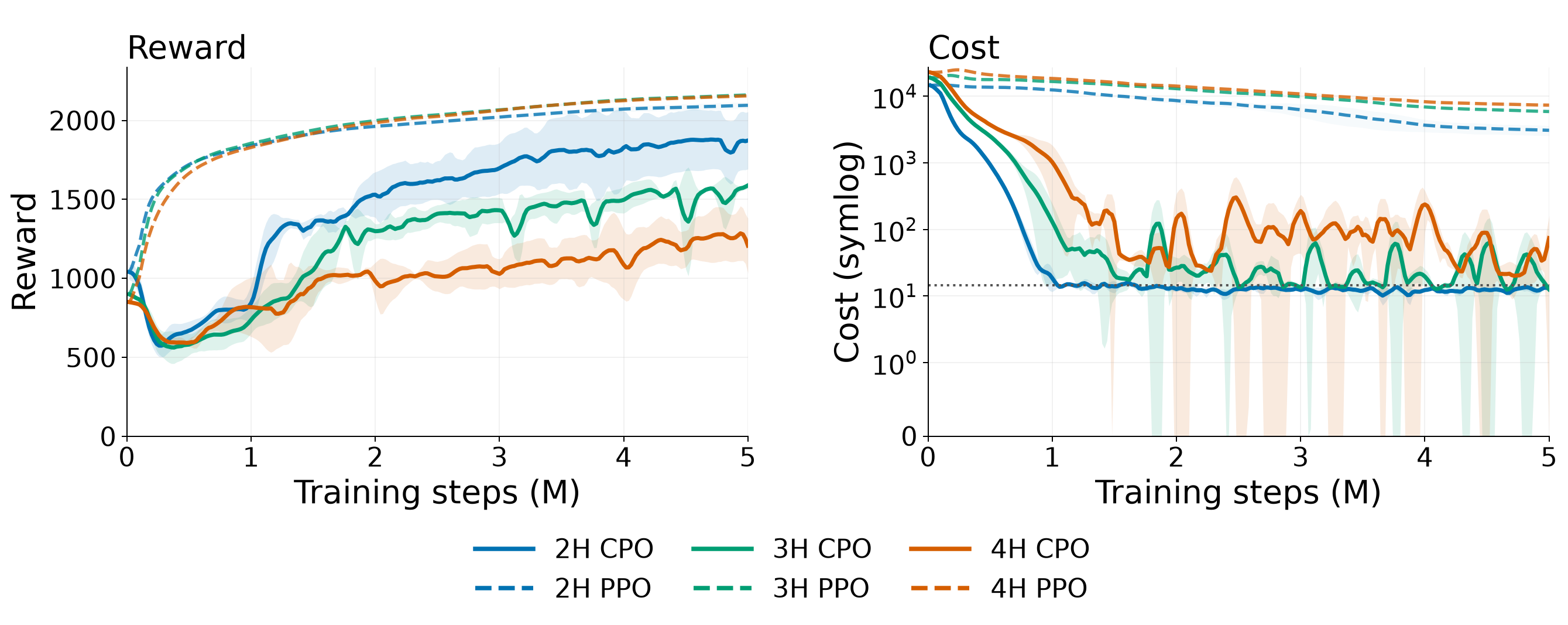}
\caption{The hand count results of CPO.}
\end{subfigure}
\hfill
\begin{subfigure}[t]{0.49\linewidth}
\centering
\includegraphics[width=\linewidth]{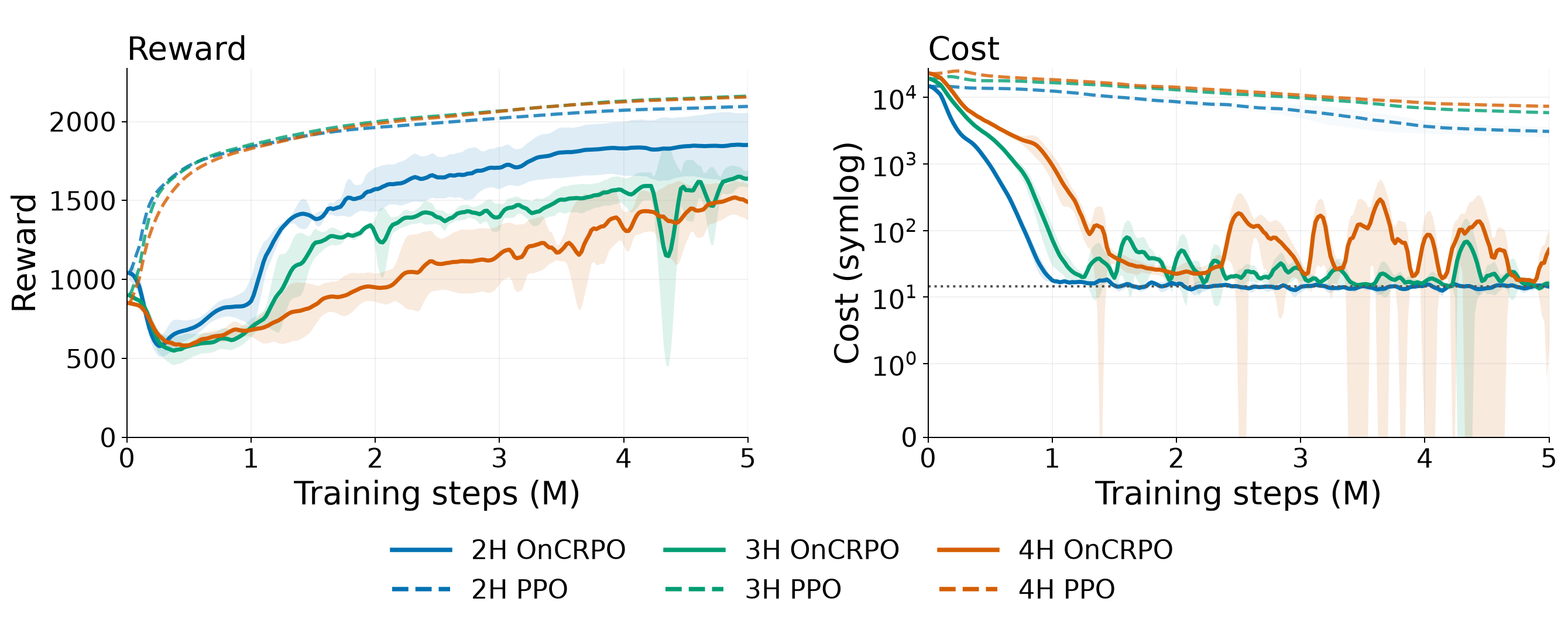}
\caption{The hand count results of OnCRPO.}
\end{subfigure}
\caption{Additional hand count results for safe RL. Solid lines denote CPO / OnCRPO; dashed lines denote PPO. Training reward and cost show means with $\pm 1$ standard deviation across three seeds. The dotted line marks the cost budget.}
\label{fig:safety-hands-additional}
\end{figure}

Figure~\ref{fig:safety-hands-additional} extends the CUP and PPO-Lag results in the main text with CPO and OnCRPO. We compare two-, three-, and four-hand configurations on the same piece, Pictures at an Exhibition: The Great Gate of Kiev, using the same actuator-power excess cost and episodic budget of $14.4$. The aggregate power allowance is fixed rather than scaled with the number of hands. Each algorithm, including the unconstrained PPO reference, is trained for 5M steps with three seeds.

Both methods substantially reduce training cost relative to PPO. However, the four-hand configuration achieves lower reward and exhibits more persistent cost fluctuations than the two-hand configuration. These results show that additional hands do not necessarily facilitate safe policy learning under a shared aggregate constraint. Because changing the hand count also changes the available interactions and control space, this comparison does not isolate action dimensionality alone. 

\subsubsection{Overall Performance Evaluation}
\label{safety overall performance evaluaction}
We compare PPO-Lag, OnCRPO, CPO, and CUP against unconstrained PPO on six tasks spanning two to four hands. The tasks combine different pieces with joint-range fraction, joint-range excess, or actuator-power excess costs. All methods use the OT reward and are trained for 5M environment steps. Reward and cost are compared within each task under a common cost definition and budget, rather than averaged across tasks with different cost scales.

\begin{figure}[H]
\centering
\begin{subfigure}[t]{0.49\linewidth}
\centering
\includegraphics[width=\linewidth]{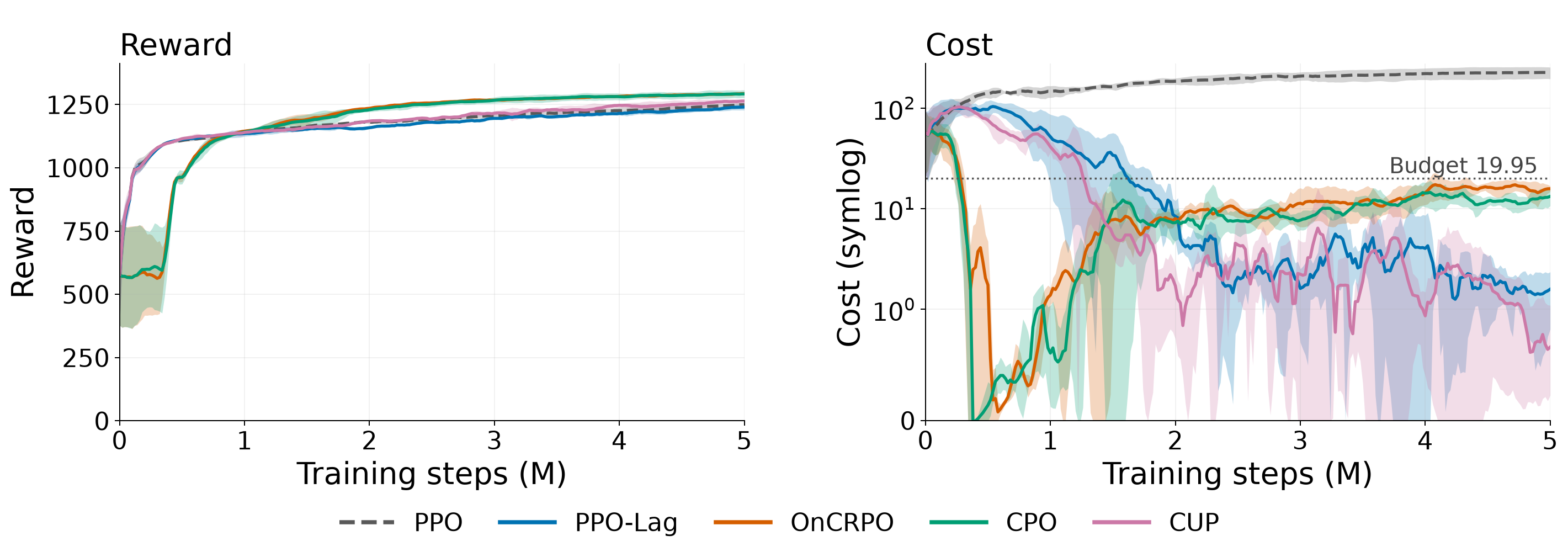}
\caption{2 hands, ForElise, joint-range fraction.}
\end{subfigure}
\hfill
\begin{subfigure}[t]{0.49\linewidth}
\centering
\includegraphics[width=\linewidth]{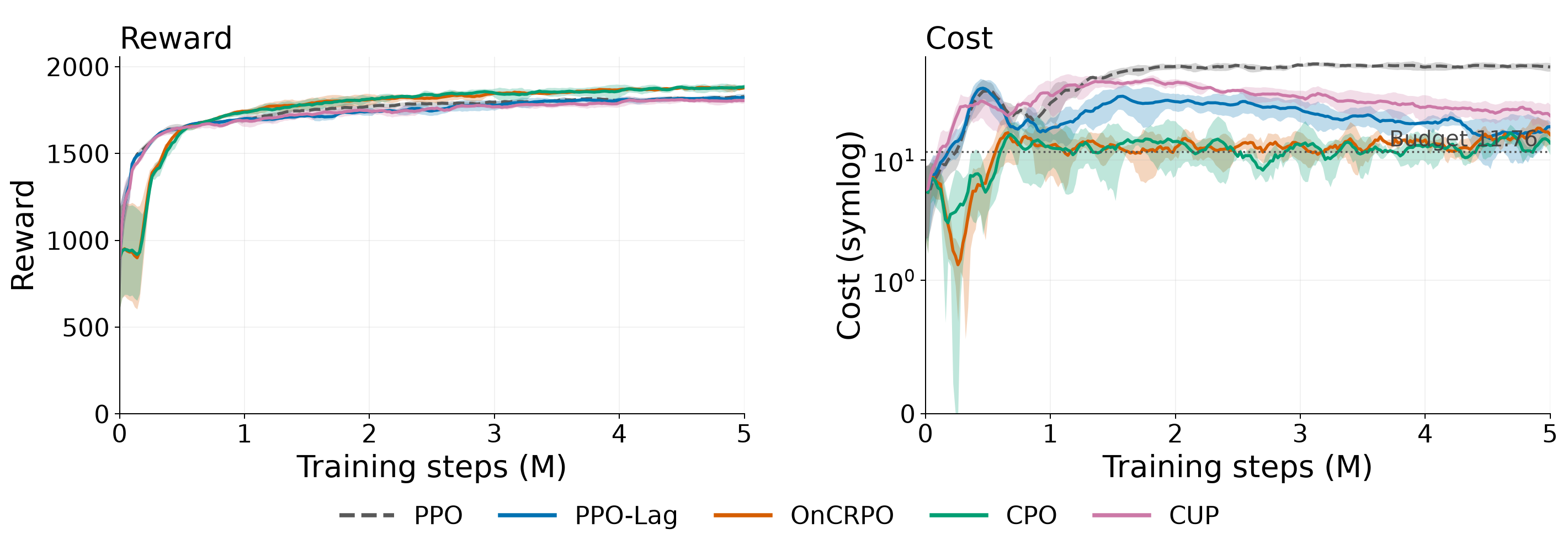}
\caption{2 hands, Clair de lune, joint-range excess.}
\end{subfigure}
\caption{Overall performance on 2 hand safety tasks: evaluation reward and cost. Shading indicates $\pm 1$ standard deviation across available seeds; dotted lines mark task-specific cost budgets.}
\label{fig:safety-overall-2h}
\end{figure}

\begin{figure}[H]
\centering
\begin{subfigure}[t]{0.49\linewidth}
\centering
\includegraphics[width=\linewidth]{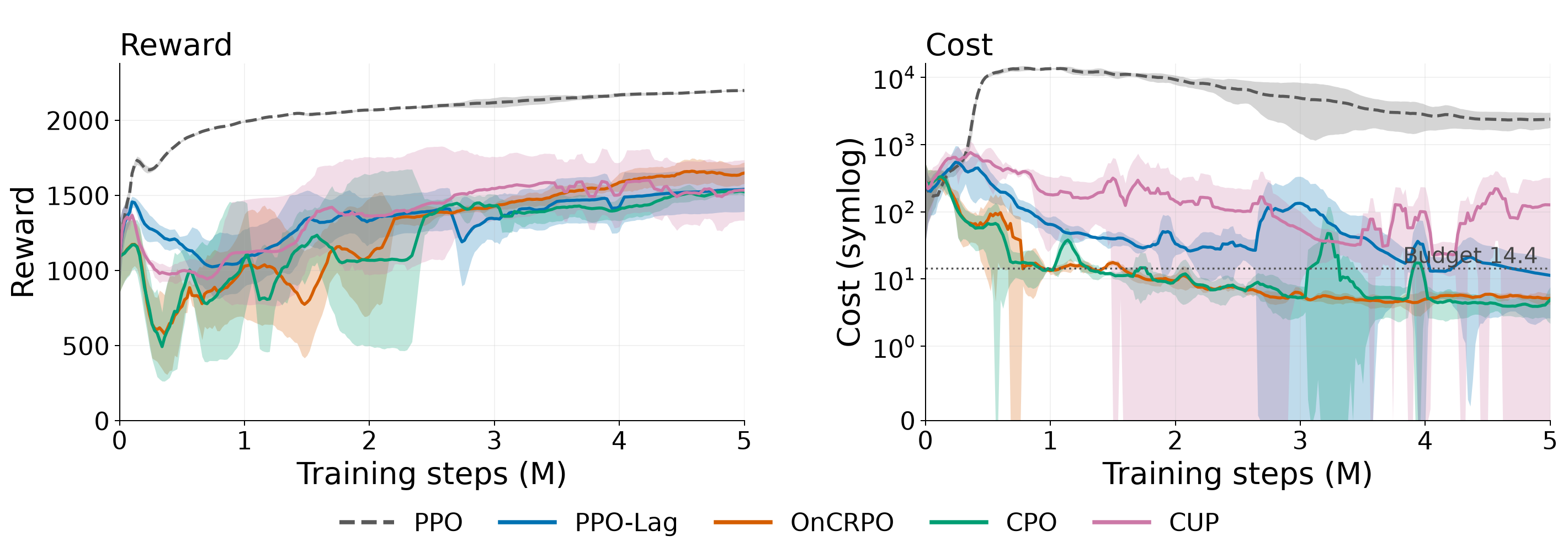}
\caption{3 hands, the Great Gate of Kiev, actuator-power excess.}
\end{subfigure}
\hfill
\begin{subfigure}[t]{0.49\linewidth}
\centering
\includegraphics[width=\linewidth]{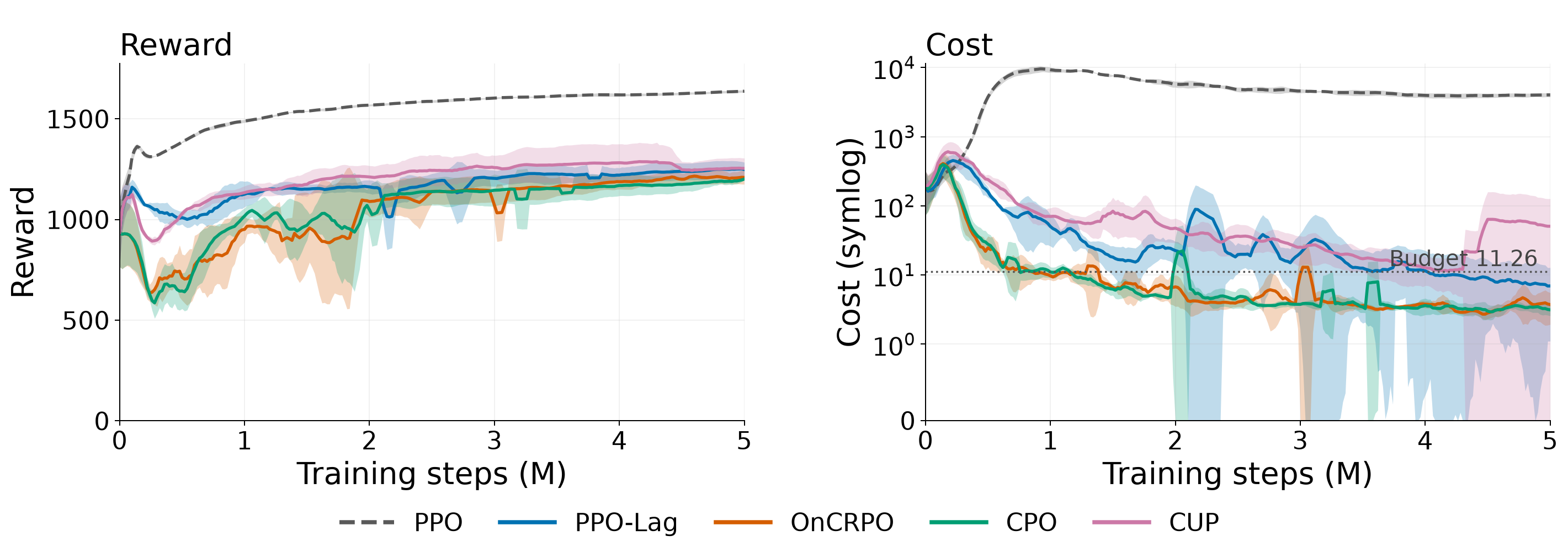}
\caption{3 hands, Polonaise Op.~40 No.~1, actuator-power excess.}
\end{subfigure}
\caption{Overall performance on 3 hand safety tasks: evaluation reward and cost. Shading indicates $\pm 1$ standard deviation across available seeds; dotted lines mark task-specific cost budgets.}
\label{fig:safety-overall-3h}
\end{figure}

\begin{figure}[H]
\centering
\begin{subfigure}[t]{0.49\linewidth}
\centering
\includegraphics[width=\linewidth]{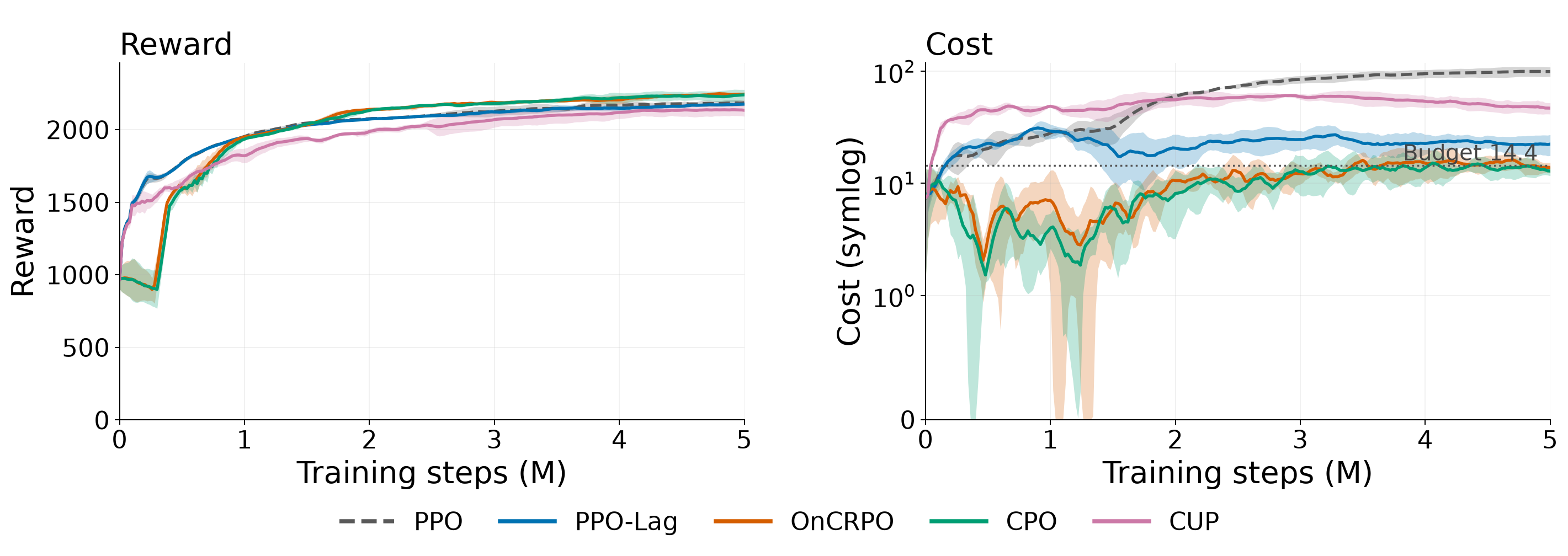}
\caption{4 hands, the Great Gate of Kiev, joint-range excess.}
\end{subfigure}
\hfill
\begin{subfigure}[t]{0.49\linewidth}
\centering
\includegraphics[width=\linewidth]{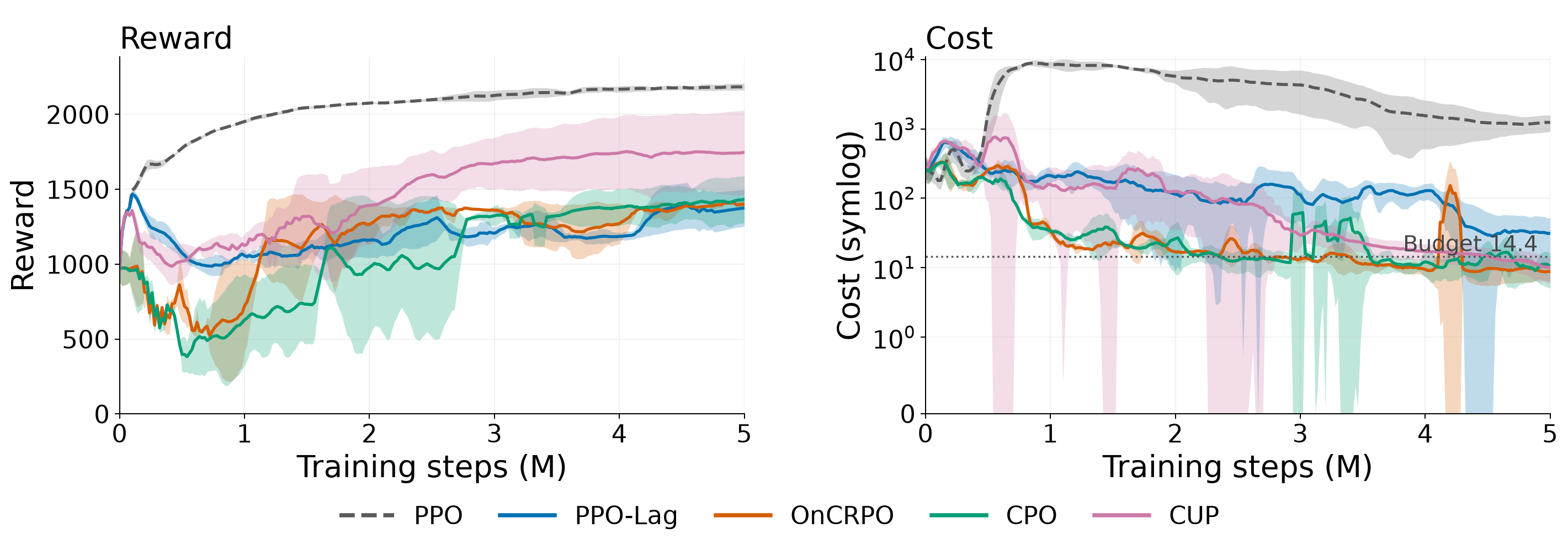}
\caption{4 hands, the Great Gate of Kiev, actuator-power excess.}
\end{subfigure}
\caption{Overall performance on 4 hand safety tasks: evaluation reward and cost. Shading indicates $\pm 1$ standard deviation across available seeds; dotted lines mark task-specific cost budgets.}
\label{fig:safety-overall-4h}
\end{figure}

\begin{figure}[H]
\centering
\begin{subfigure}[t]{0.49\linewidth}
\centering
\includegraphics[width=\linewidth]{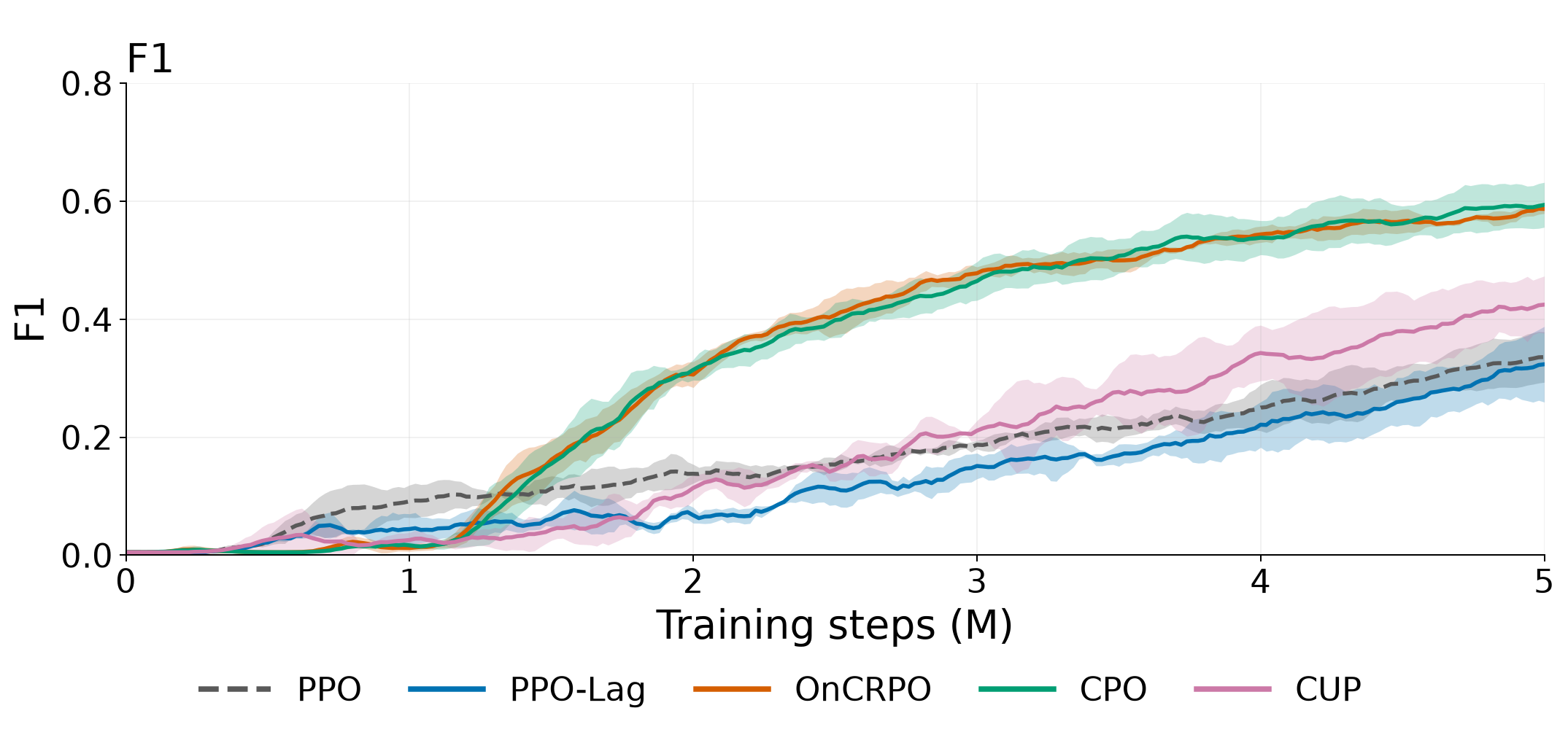}
\caption{2 hands, ForElise, joint-range fraction.}
\end{subfigure}
\hfill
\begin{subfigure}[t]{0.49\linewidth}
\centering
\includegraphics[width=\linewidth]{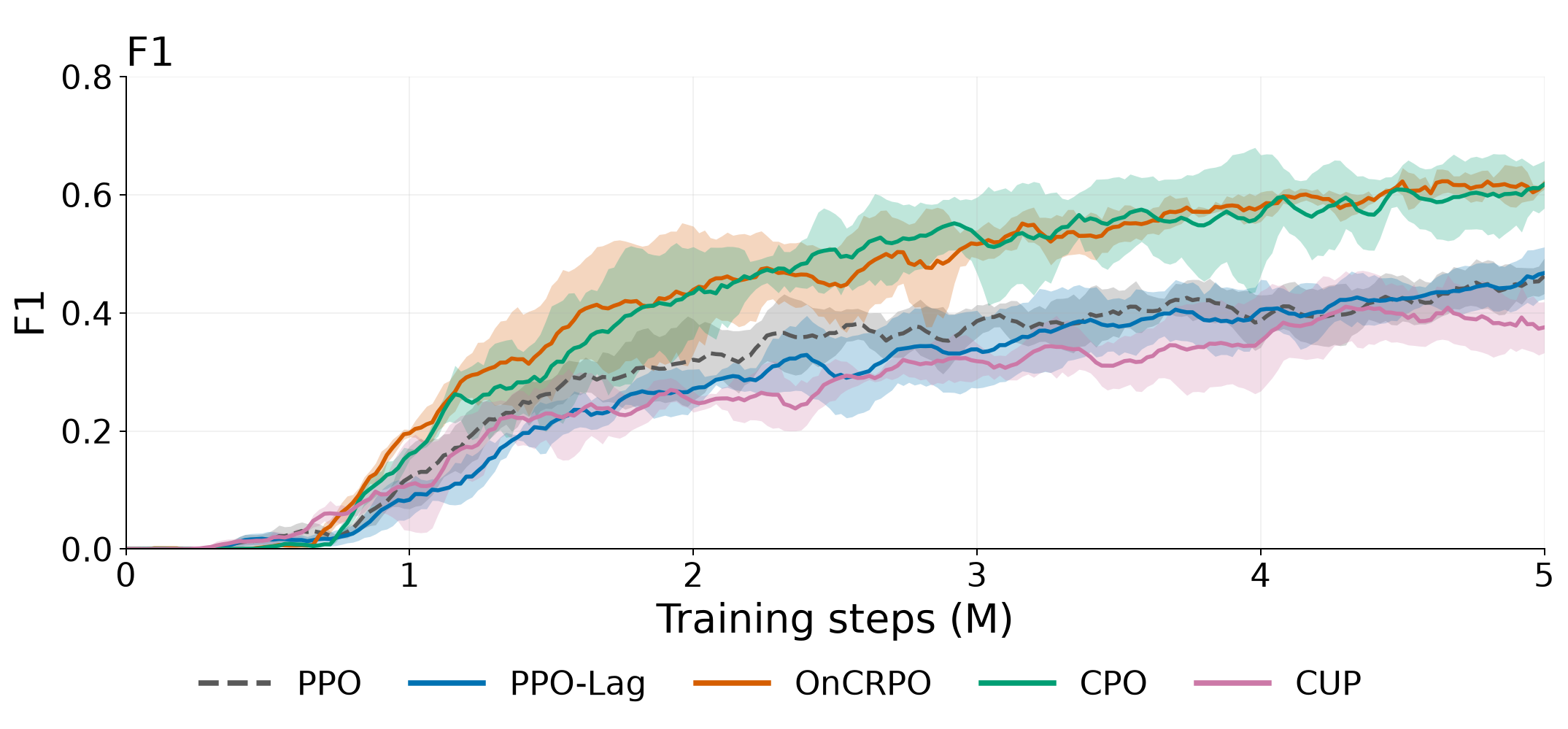}
\caption{2 hands, Clair de lune, joint-range excess.}
\end{subfigure}
\caption{Key-press F1 on the 2-hand safety tasks. Curves show means with $\pm 1$ standard deviation across available seeds.}
\label{fig:safety-f1-2h}
\end{figure}

\begin{figure}[H]
\centering
\begin{subfigure}[t]{0.49\linewidth}
\centering
\includegraphics[width=\linewidth]{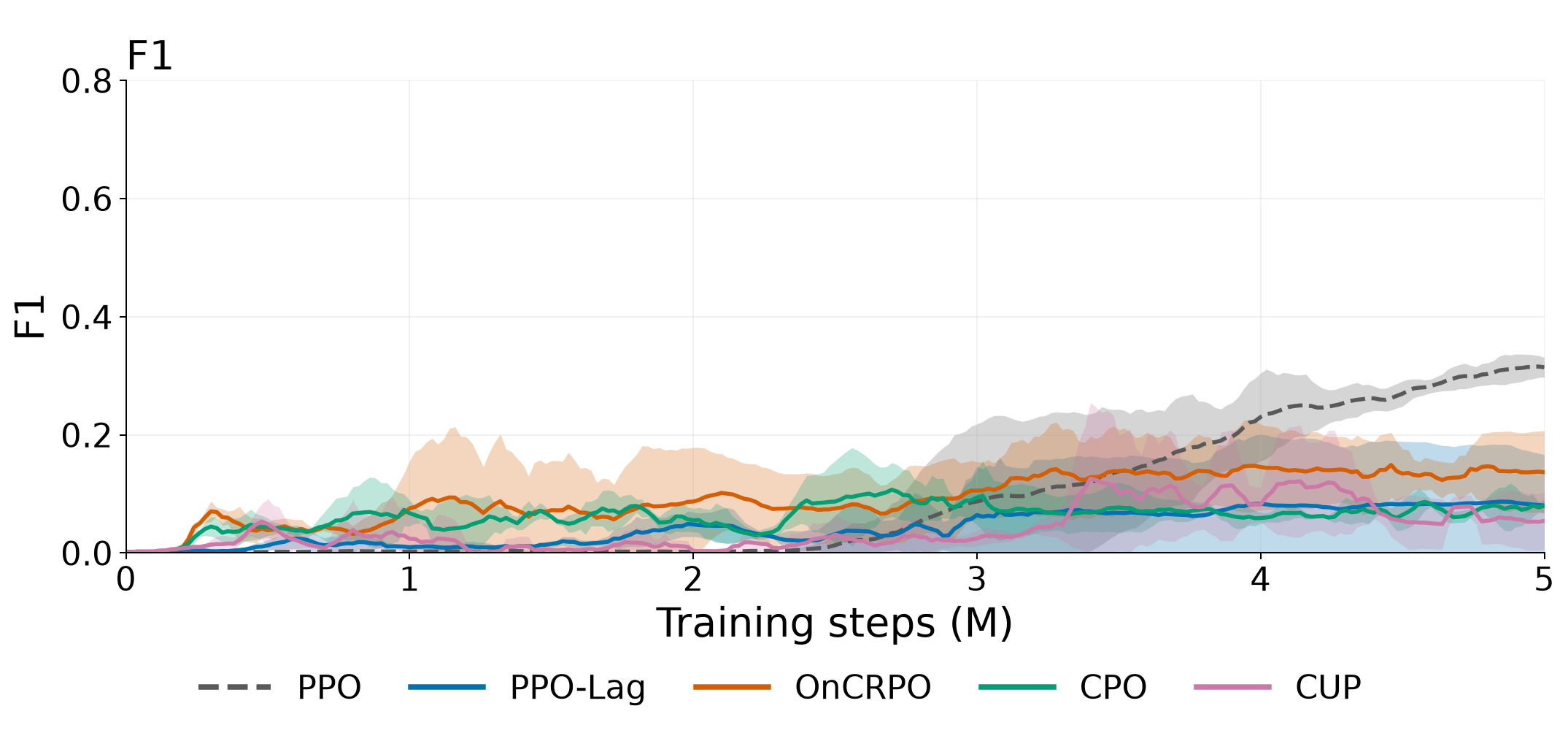}
\caption{3 hands, the Great Gate of Kiev, actuator-power excess.}
\end{subfigure}
\hfill
\begin{subfigure}[t]{0.49\linewidth}
\centering
\includegraphics[width=\linewidth]{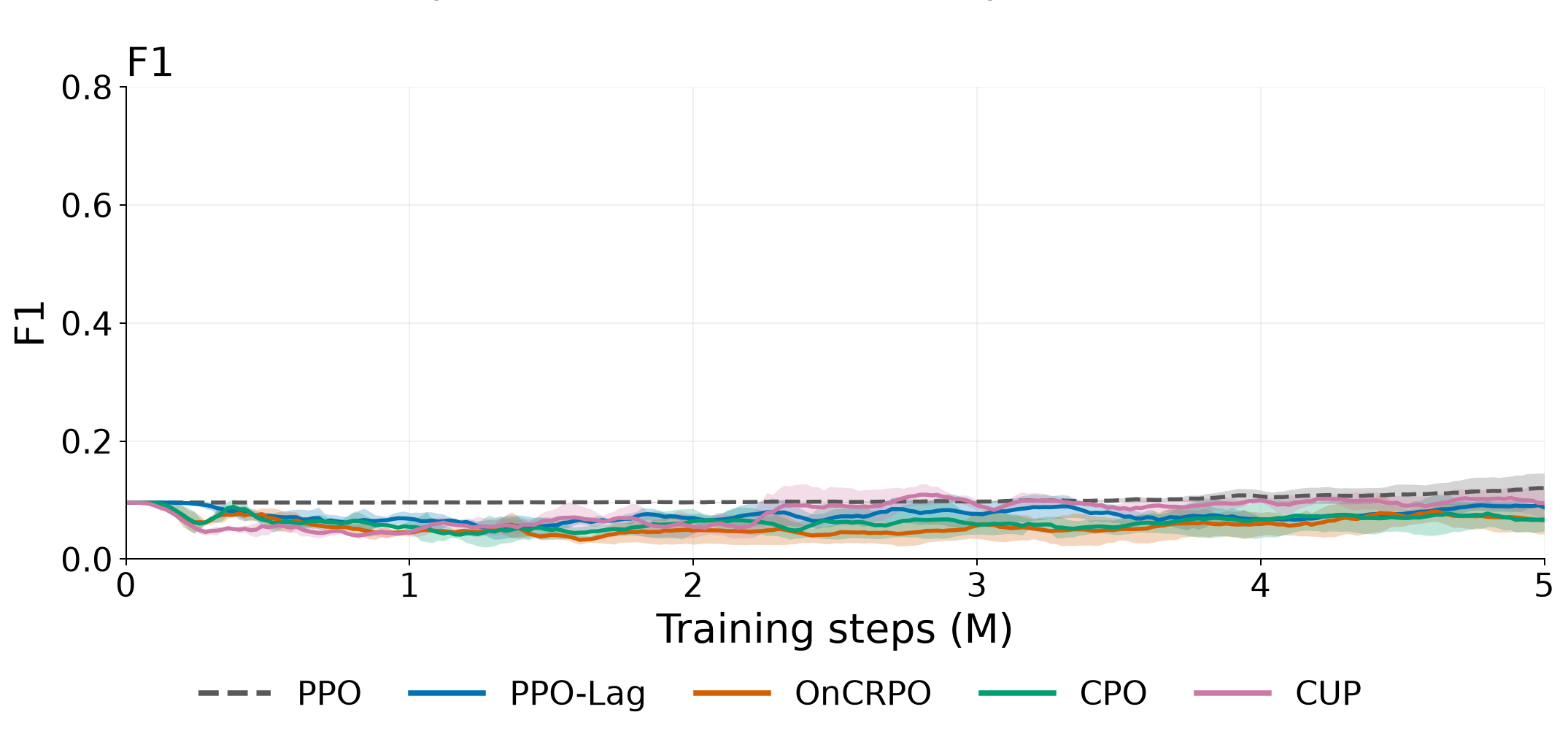}
\caption{3 hands, Polonaise Op.~40 No.~1, actuator-power excess.}
\end{subfigure}
\caption{Key-press F1 on the 3-hand safety tasks. Curves show means with $\pm 1$ standard deviation across available seeds.}
\label{fig:safety-f1-3h}
\end{figure}

\begin{figure}[H]
\centering
\begin{subfigure}[t]{0.45\linewidth}
\centering
\includegraphics[width=\linewidth]{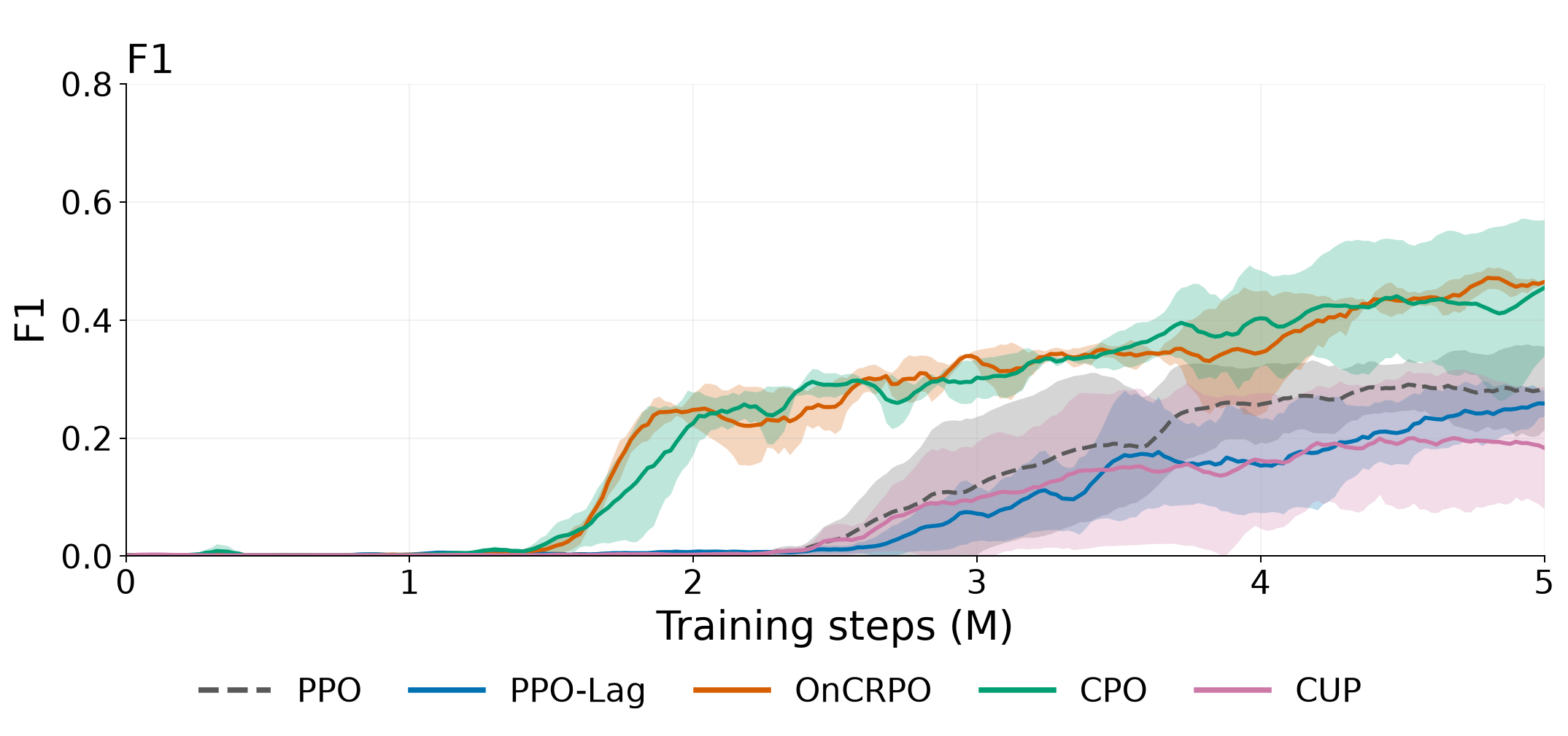}
\caption{4 hands, the Great Gate of Kiev, joint-range excess.}
\end{subfigure}
\hfill
\begin{subfigure}[t]{0.45\linewidth}
\centering
\includegraphics[width=\linewidth]{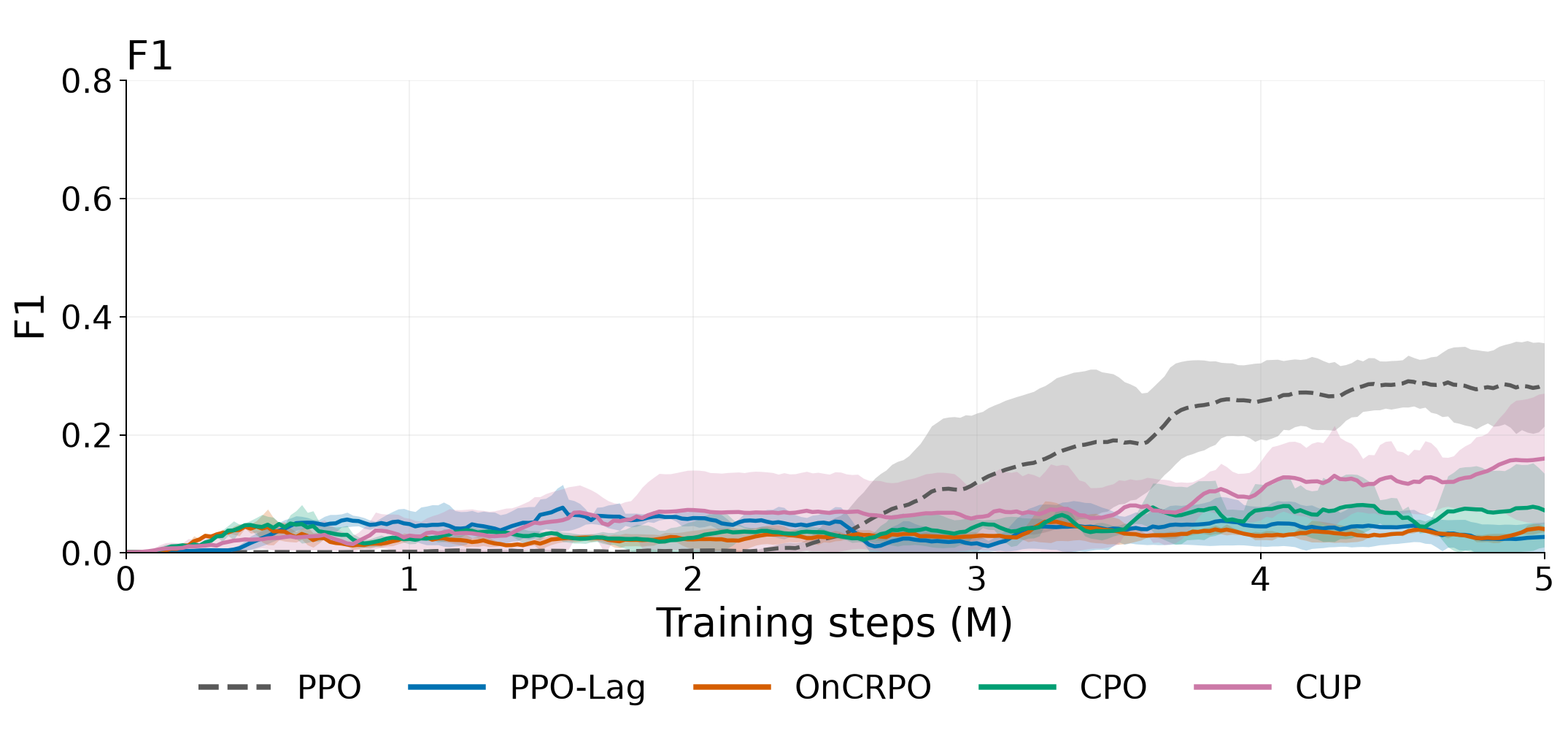}
\caption{4 hands, the Great Gate of Kiev, actuator-power excess.}
\end{subfigure}
\caption{Key-press F1 on the 4-hand safety tasks. Curves show means with $\pm 1$ standard deviation across available seeds.}
\label{fig:safety-f1-4h}
\end{figure}

Figures~\ref{fig:safety-overall-2h}--\ref{fig:safety-overall-4h} report checkpoint-evaluation reward and undiscounted episodic cost. The corresponding key-press F1 curves appear in Figures~\ref{fig:safety-f1-2h}--\ref{fig:safety-f1-4h}. Reporting all three metrics distinguishes cost reduction from successful piano playing: lower cost alone does not establish that a policy preserves musical performance.

Curves use the available completed evaluation runs. Each task--algorithm group contains three seeds. For trajectory plots, we apply a trailing five-point average to each seed before computing the mean and sample standard deviation. The algorithms learn to reduce safety costs across the six tasks, but their playing performance depends on the constraint. Under joint range constraints, CPO and OnCRPO reduce cost without sacrificing reward or F1 relative to PPO. Under power constraints, the algorithms also reduce cost, but achieve lower reward and F1. Therefore, adding safety constraints can make learning more difficult, as the agent must learn to play the piano while also reducing safety costs.

\subsubsection{Sensitivity to the Cost Limit}
\label{cost list section}
We vary the cost limit while keeping the piece,
hand configuration, safety semantic, and cost definition fixed.
A lower limit imposes a stricter safety requirement by allowing
less cumulative cost per episode.
We consider four limits, $d\in\{4.32,14.4,43.2,144\}$,
on the actuator power task with four hands.
CPO and CUP are each trained for 5M steps with three seeds
per limit.

\begin{figure}[H]
\centering
\begin{subfigure}[t]{0.49\linewidth}
\centering
\includegraphics[width=\linewidth]{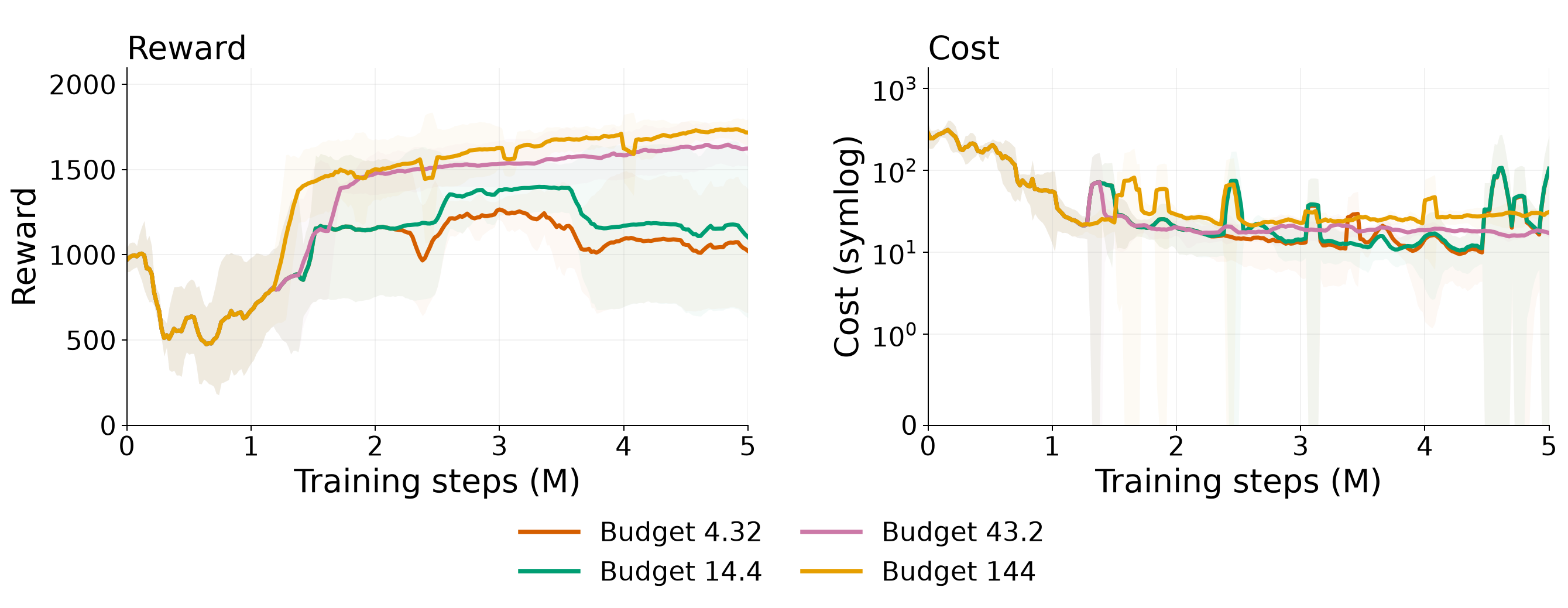}
\caption{The budget sensitivity results of CPO.}
\end{subfigure}
\hfill
\begin{subfigure}[t]{0.49\linewidth}
\centering
\includegraphics[width=\linewidth]{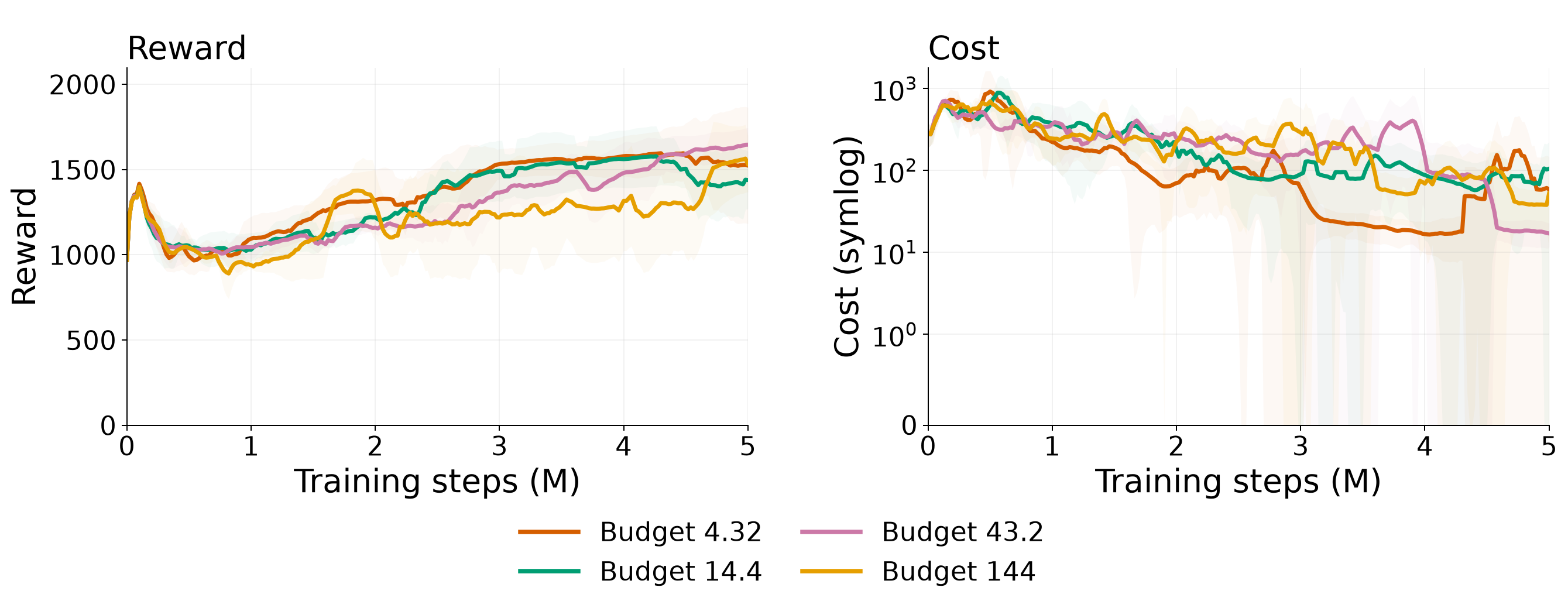}
\caption{The budget sensitivity results of CUP.}
\end{subfigure}
\caption{Sensitivity to the episodic cost limit. Evaluation reward and cost are shown for five budgets; shading indicates $\pm 1$ standard deviation across three seeds.}
\label{fig:safety-budget}
\end{figure}

\begin{figure}[H]
\centering
\begin{subfigure}[t]{0.49\linewidth}
\centering
\includegraphics[width=\linewidth]{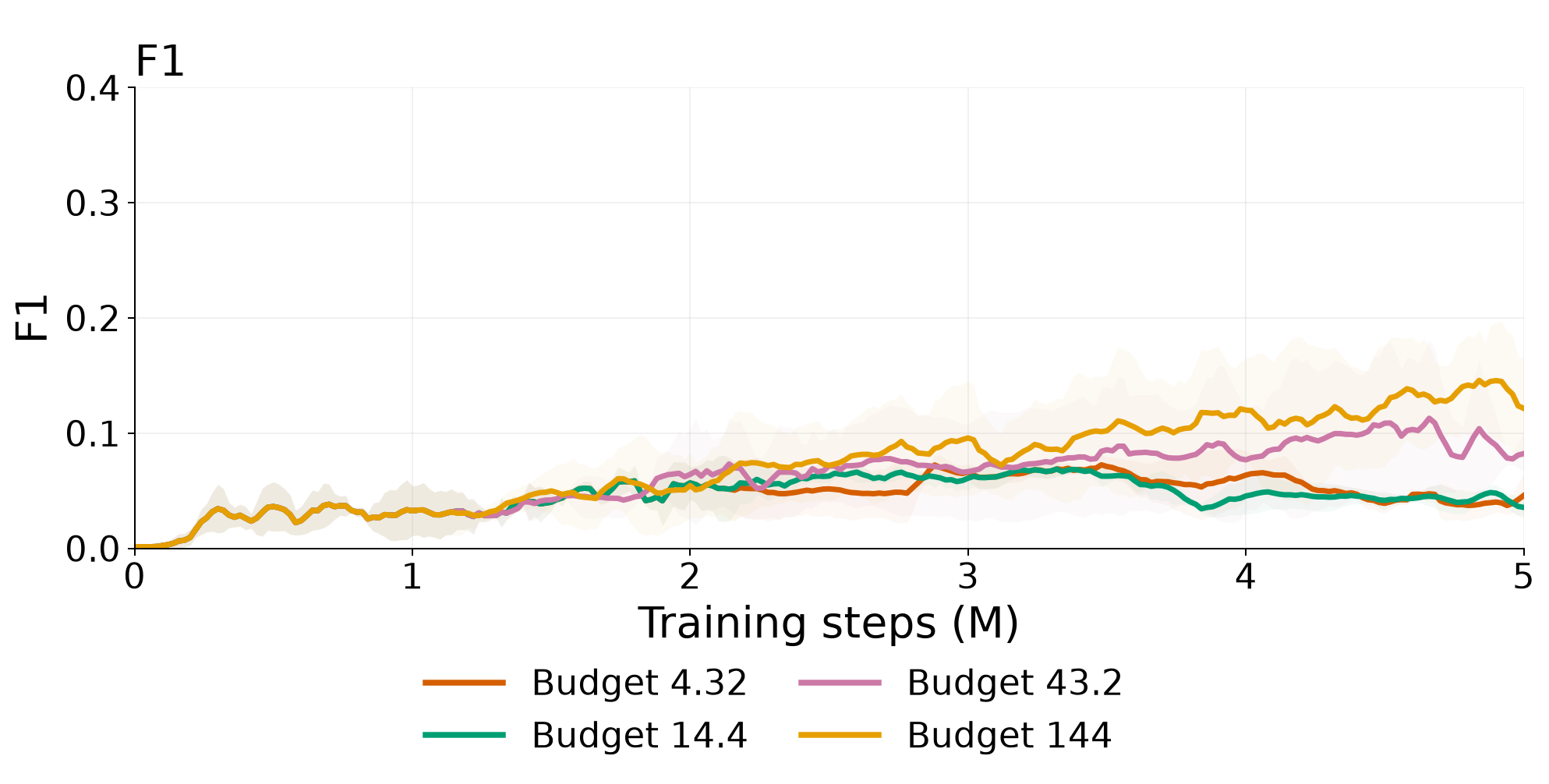}
\caption{The budget sensitivity F1 of CPO.}
\end{subfigure}
\hfill
\begin{subfigure}[t]{0.49\linewidth}
\centering
\includegraphics[width=\linewidth]{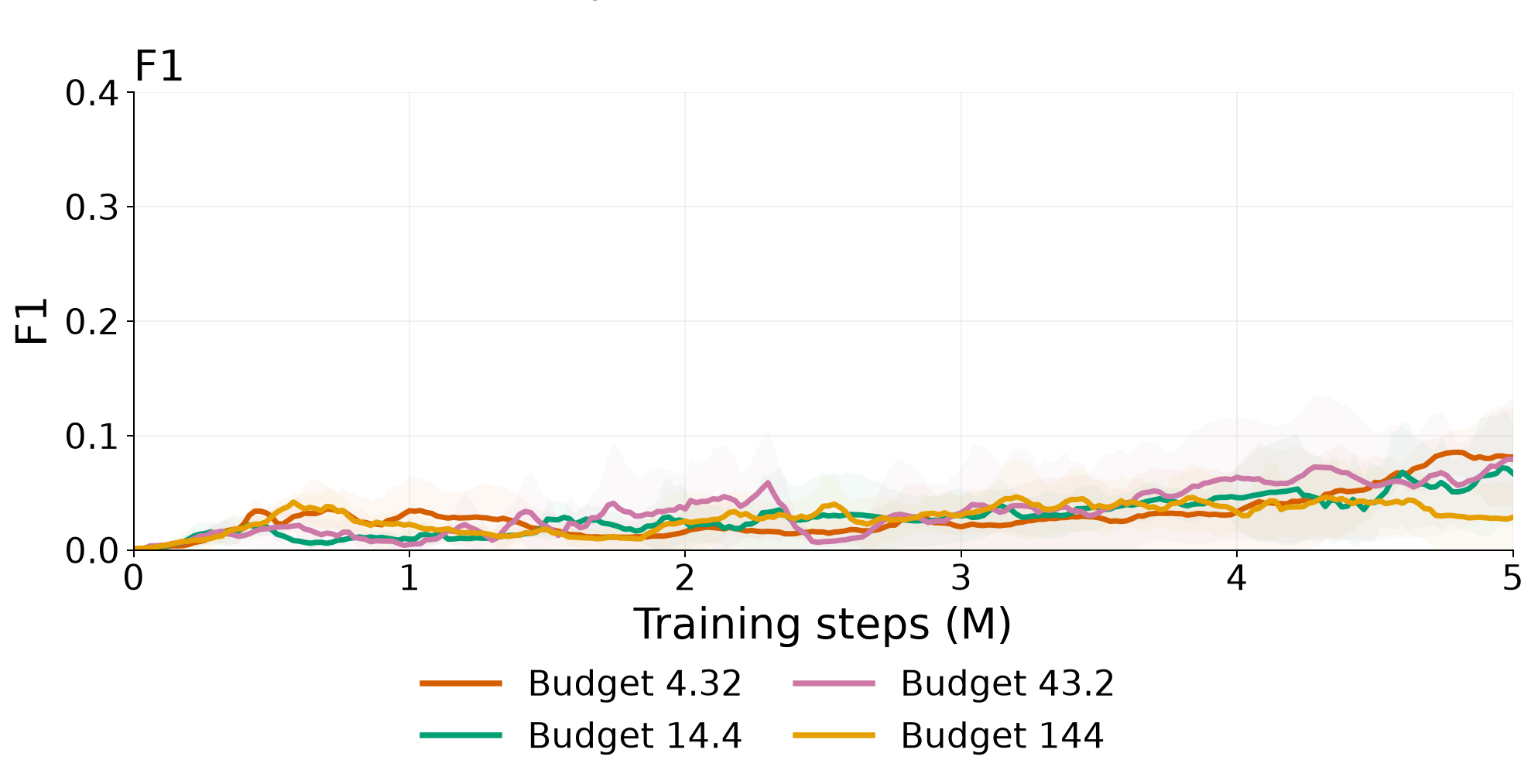}
\caption{The budget sensitivity F1 of CUP.}
\end{subfigure}
\caption{Key-press F1 under different cost limits, shown as means with $\pm 1$ standard deviation across three seeds.}
\label{fig:safety-budget-f1}
\end{figure}

Figures~\ref{fig:safety-budget} and~\ref{fig:safety-budget-f1}
show how these limits affect reward, cost, and F1.
For CPO, the stricter limits of $4.32$ and $14.4$ produce
lower reward and F1 than the more permissive limits of
$43.2$ and $144$.
This suggests that stricter safety requirements can make
accurate piano playing harder to learn.
For CUP, however, relaxing the limit does not consistently
improve performance: the largest limit produces the lowest
final mean F1.
Both algorithms also finish training with mean costs above
the strictest limit of $4.32$, indicating that this requirement
remains difficult to satisfy within 5M steps.

\FloatBarrier
\subsection{Multi-Agent RL}
\label{app:marl_exp_result}

\begin{table}[H]
\centering
\caption{MARL test settings.}
\label{tab:settings}

\small
\renewcommand{\arraystretch}{0.94}
\begin{tabular*}{\linewidth}{@{\extracolsep{\fill}}lccccc@{}}
\toprule
Setting & Agent & Hands & Action & Observation & Visible hands \\
\midrule

\multirow{2}{*}{Base}
& A1 & 0,1 & 1--44  & 1--49  & 2 \\
& A2 & 2,3 & 45--88 & 40--88 & 1 \\
\specialrule{0.3pt}{1.5pt}{1.5pt}

\multirow{2}{*}{Observability}
& A1 & 0,1 & 1--44  & 1--70  & 2 \\
& A2 & 2,3 & 45--88 & 19--88 & 1 \\
\specialrule{0.3pt}{1.5pt}{1.5pt}

\multirow{2}{*}{Coupling}
& A1 & 0,1 & 1--70  & 1--49  & 2 \\
& A2 & 2,3 & 19--88 & 40--88 & 1 \\
\specialrule{0.3pt}{1.5pt}{1.5pt}

\multirow{2}{*}{Heterogeneity}
& A1 & 0     & 1--22  & 1--27  & 1 \\
& A2 & 1,2,3 & 23--88 & 18--88 & 0 \\
\specialrule{0.3pt}{1.5pt}{1.5pt}

\multirow{4}{*}{Scalability}
& A1 & 0 & 1--22  & 1--27  & 1 \\
& A2 & 1 & 23--44 & 18--49 & 0,2 \\
& A3 & 2 & 45--66 & 40--71 & 1,3 \\
& A4 & 3 & 67--88 & 62--88 & 2 \\
\bottomrule
\end{tabular*}
\end{table}

This section provides detail MARL experiment setup , results and analysis. Table~\ref{tab:settings} summarizes the five MARL task configurations. The \textbf{Base task} uses two agents, each controlling two hands and a disjoint 44-key action range, with only 10-key overlap between their local observations. The \textbf{Observability Variant} keeps the same hand and action assignments but broadens the observation ranges from 49 keys to 70 keys, so that both agents observe a much larger shared keyboard region. The \textbf{Coupling Variant} instead keeps the Base observations while expanding the action ranges to 1--70 and 19--88, creating a 52-key overlap that both agents can control. The \textbf{Heterogeneity Variant} assigns one hand and 22 actions to the first agent, but three hands and 66 actions to the second, producing a strongly imbalanced control load. The \textbf{Scalability Variant} divides the same four hands among four agents, each controlling one hand and 22 actions, with locally centered observation ranges that overlap by 10 keys between neighboring agents. During training, we additionally discourage physical interference between agents with an inter-agent collision penalty. At each control step, if any physical contact occurs between hand parts controlled by different agents, we subtract $0.1$ once from the shared team reward, regardless of the number or force of contacts. The resulting training reward is
\[
r_t^{\mathrm{train}} = r_t - \lambda_c c_t^{\mathrm{inter}},
\]
where $r_t$ is the original benchmark reward in Equation~\ref{eq:ot reward}, $c_t^{\mathrm{inter}}$ indicates inter-agent contact, and $\lambda_c$ controls the penalty strength and is set to $0.1$ in our experiments. The same shaped reward is shared by all agents, while evaluation uses the original benchmark reward.

\begin{figure}[H]
    \centering
    \includegraphics[width=\textwidth]
    {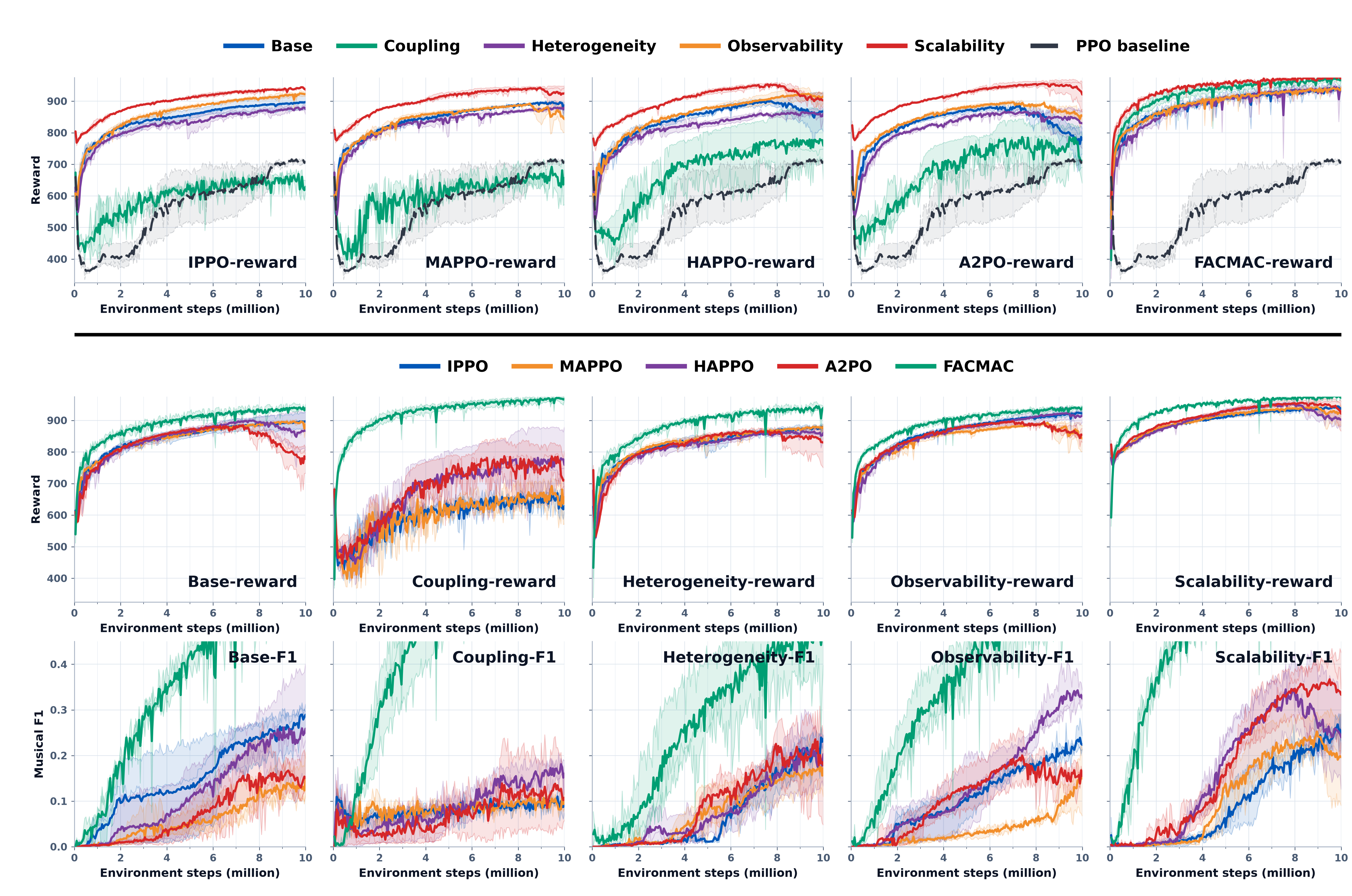}
    \caption{
    Detailed MARL results across the five \ourbenchmark cooperation settings.
    The first row compares task-wise team return for each MARL algorithm, while the second and third rows compare team return and musical F1 across algorithms for each task.
    }
    \label{fig:marl_15_panel_comparison}
\end{figure}

The first row of Figure~\ref{fig:marl_15_panel_comparison} compares how each MARL algorithm responds to the five task configurations. Across algorithms, the setting with \textbf{Increased Scalability} consistently achieves the highest or near-highest return. The physical four-hand task and total action space remain unchanged, but control is divided among four agents instead of two. Compared with Base, this reduces each agent's 

assignment from two hands and 44 actions to one hand and 22 actions, while increasing the number of independently acting controllers from two to four. The higher return suggests that the benefit from reduced per-agent control burden appears to outweigh the drawback brought by additional coordination. \textbf{Increased Observability} also generally improves or matches Base, showing that broader actor observations at execution time can improve coordination by exposing more keyboard and teammate information. In contrast, the \textbf{Heterogeneity} setting usually reduces performance, possibly because one agent must control three hands and 66 actions, making its local control problem much harder. The setting with \textbf{Increased Coupling} produces the largest degradation for A2PO, HAPPO, IPPO, and MAPPO, which combines overlapping control with severe partial observability. FACMAC is the clear exception and remains strong under Coupling, possibly because its centralized factored critic and joint policy gradient better capture the effects of interacting agent actions.

The second and third rows of Figure~\ref{fig:marl_15_panel_comparison} summarize algorithm performance across the five MARL tasks using team return and musical F1. Because the return curves are close for several methods, we additionally report musical F1 following RoboPianist~\citep{zakka2023robopianist}, which is more sensitve to algorithms' performance and more directly measures note execution accuracy. At each timestep, precision measures how many pressed keys are correct, recall measures how many target keys are successfully pressed, and F1 is their harmonic mean.

Across the five tasks, \textbf{Coupling} provides the clearest separation between algorithms. The two agents share a 52-key action overlap, while 21 keys in each agent's action range lie outside its own observation range, combining strong cross-agent action interference with severe partial observability. Under this setting, HAPPO and A2PO achieve higher returns than IPPO and MAPPO. FACMAC performs best overall on this task, consistent with the sample efficiency of its off-policy training. 

Similar differences persist across the other tasks as well, especially in F1 even when returns are close. IPPO remains a surprisingly strong baseline, matching or outperforming MAPPO in most settings and often achieving higher F1, indicating that MAPPO's centralized critic does not consistently improve performance in \ourbenchmark. One possible explanation is that its larger critic input is harder to fit, although our experiments do not isolate this effect. Since HAPPO and A2PO also 

\begin{figure}[H]
    \centering
    \includegraphics[width=0.82\linewidth]{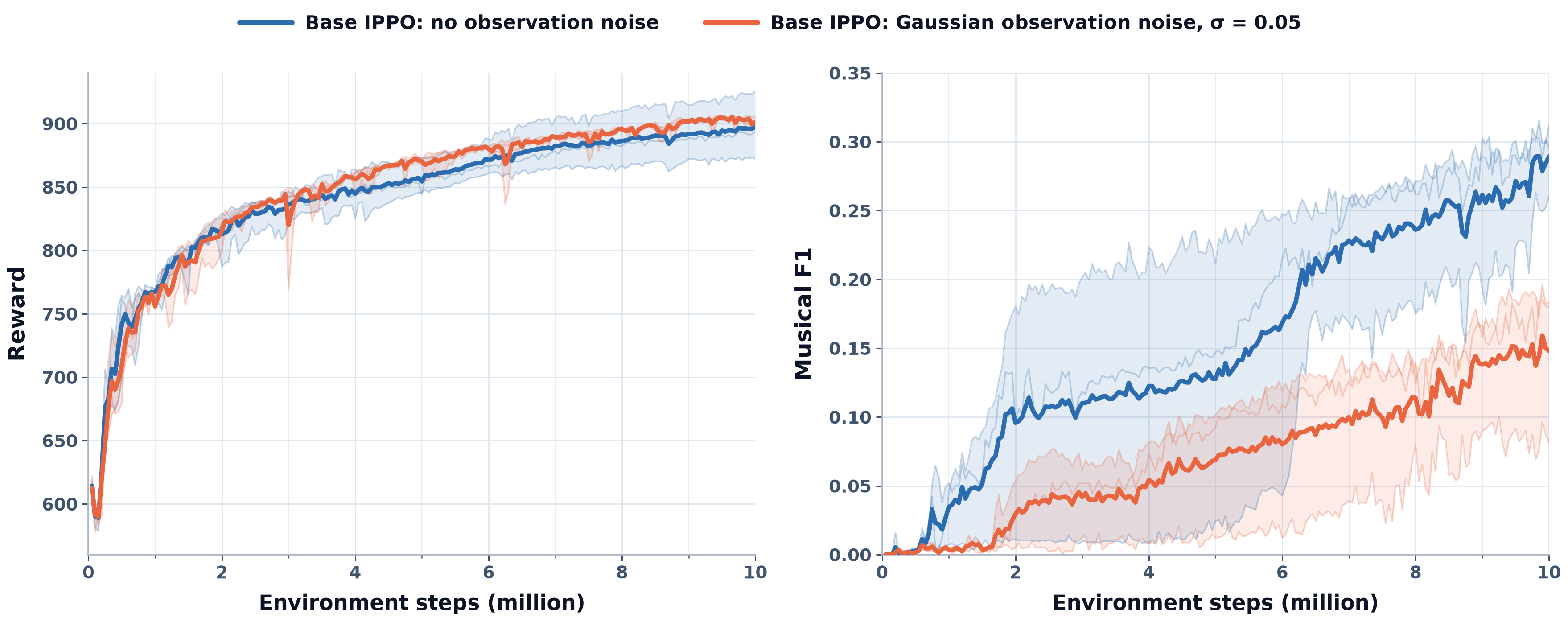}
    \caption{IPPO performance in the Base MARL setting with and without Gaussian observation noise ($\sigma=0.05$). Reward remains similar, while musical F1 degrades substantially under observation noise.}
    \label{fig:marl_obs_noise}
\end{figure}

use global critics without showing the same weakness, centralized value information may be most useful when combined with policy updates that better account for inter-agent interactions. Finally, centralized PPO performs markedly worse than the MARL methods across all tasks, supporting the benefit of factorizing this high-dimensional control problem across multiple agents.

Finally, we further test the Base IPPO setting under Gaussian observation noise with $\sigma=0.05$, applied at every control step to hand joint states, piano states, and sustain states while leaving the MIDI goal unchanged. As shown in Figure~\ref{fig:marl_obs_noise}, the reward curves remain close to the clean baseline and largely fall within normal seed variation. In contrast, musical F1 drops substantially under observation noise, reaching about 0.15 compared with roughly 0.29 without noise. This shows that observation perturbations can noticeably degrade note-execution quality even when the training return appears similar. This is consistent with the Robust RL results in Section~\ref{sec:robust_rl_results}, where policy performance also degrades under injected perturbations, with the severity depending on the perturbation channel.

\section{Detailed Experimental Settings}
\label{app:experimental_settings}
\subsection{Standard RL}
\subsubsection{Reward Setting}
\label{app:detailed_reward_setting}
The \textit{fingering-annotated reward} directly uses fingering annotations
from the MIDI score. In contrast, the \textit{optimal transport (OT) reward}
constructs a fingertip-to-key cost matrix from their spatial distances and
solves the resulting minimum-cost one-to-one assignment using the Hungarian
algorithm \citep{chen2025dexterous}. The OT formulation removes the dependence on fingering annotations
and therefore supports arbitrary MIDI files and hand configurations, whereas
the annotation-based reward provides explicit finger-level supervision but is
limited to two-hand settings.
\subsubsection{Evaluation Metric}
\label{app:evaluation_metric}

We use the F1 score to evaluate key-activation accuracy by comparing the keys played by the policy with the target keys specified by the MIDI score at each control step. A true positive (TP) denotes a target key that is correctly activated, a false positive (FP) denotes a non-target key that is incorrectly activated, and a false negative (FN) denotes a target key that is not activated. Precision therefore measures how many played keys are correct, while recall measures how many target keys are successfully played. The F1 score is computed from precision and recall as
    \begin{equation}
    \label{eq:f1}
    \begin{aligned}
    \mathrm{Precision} &= \frac{\mathrm{TP}}{\mathrm{TP}+\mathrm{FP}}, \qquad
    \mathrm{Recall} = \frac{\mathrm{TP}}{\mathrm{TP}+\mathrm{FN}}, \\
    \mathrm{F1} &= \frac{2\,\mathrm{Precision}\,\mathrm{Recall}}
    {\mathrm{Precision}+\mathrm{Recall}}.
    \end{aligned}
    \end{equation}
\begin{figure}[H]
    \centering
    \includegraphics[width=1\linewidth]{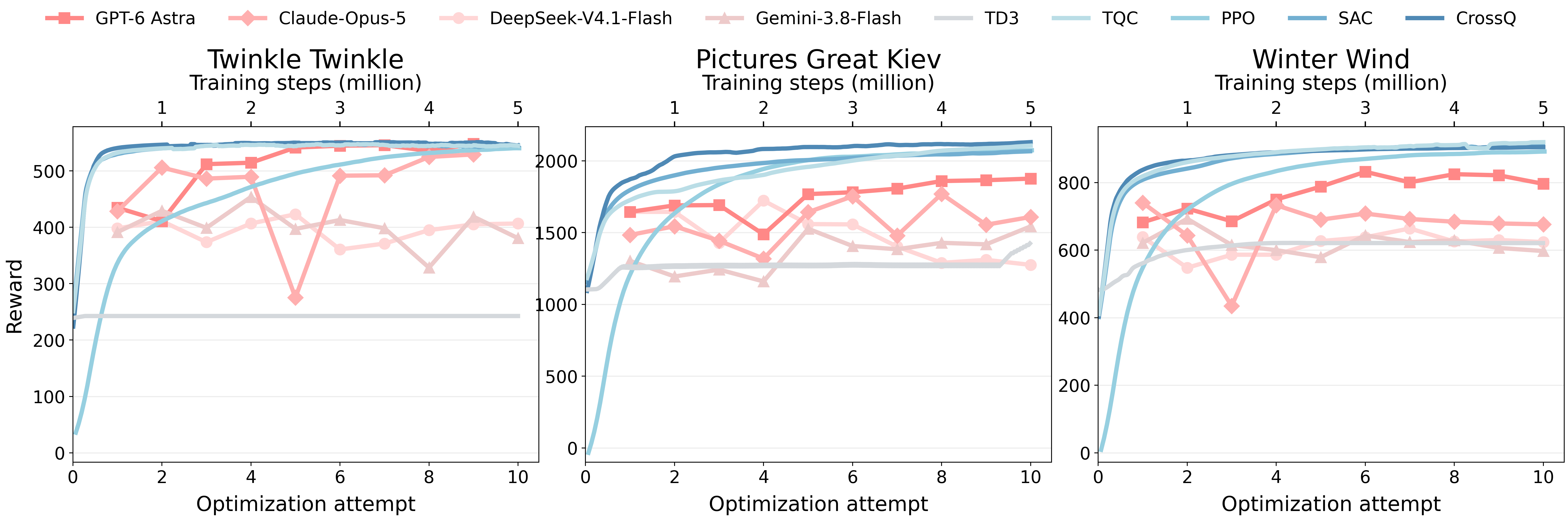}
    \caption{Rollout reward of llms over ten keyframe-optimization iterations and RL method over 5M steps for the two-hand setting on three piano pieces.}
    \label{fig:twohand_final_eval_reward}
\end{figure}

The F1 score balances the two complementary error modes: pressing incorrect keys and missing required keys. A high F1 score therefore requires both accurate and complete execution of the MIDI targets, making it a direct measure of task performance in our piano-playing setting.

\subsubsection{LLM Setting}
\label{app:detailed_llm}
\begin{figure}[H]
    \centering
    \includegraphics[width=\linewidth]{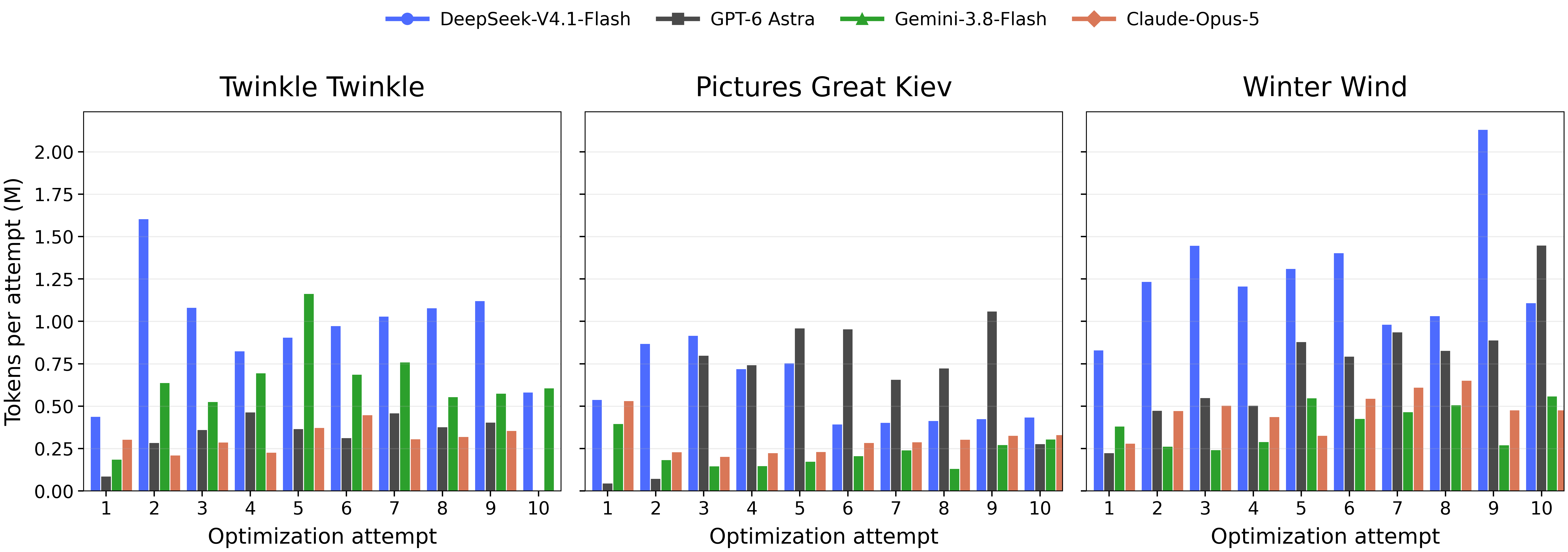}
    \caption{token consumption over ten keyframe-optimization iterations for the two-hand setting on three piano pieces.}
    \label{fig:general-twohand-f1-token-efficiency}
\end{figure}

\begin{description}[
    leftmargin=0pt,
    labelindent=0pt,
    itemsep=1pt,
    parsep=0pt,
    topsep=2pt,
    labelsep=0.5em
]
    \item[\textbf{LLM-based agent framework.}] Our agent follows the code--feedback--optimization paradigm, adapted to dexterous piano playing with Shadow Hands. Rather than directly predicting low-level actions, the LLM synthesizes a complete Python controller that generates an open-loop actuator trajectory for each performance. The controller is executed in simulation and evaluated using task-level metrics, including F1 score, reward, note-timing errors, and trajectory diagnostics. These results are appended to the conversation as feedback, enabling iterative in-context refinement of the controller.

    \item[\textbf{Tool configuration.}] Following \citet{zhang2026unexpectedrobotpolicyearly}, we provide the agent with a forward-kinematics tool for inspecting candidate hand poses. Beyond this, we introduce several task-specific tools for dexterous piano playing. The agent receives structured information about the MIDI score and timing, hand configuration, actuator bounds, and robot geometry, and can query Shadow-Hand reference information and retrieve local trajectory segments around erroneous time steps. These additions expose task structure and execution diagnostics to the LLM, enabling more targeted controller refinement from simulator feedback.

    \item[\textbf{Optimization protocol.}] For each piano piece, the agent performs ten controller-optimization attempts. Each attempt uses feedback from previous executions to revise the controller, while model parameters remain fixed. We report the controller achieving the highest F1 score across the ten attempts.The reward results and F1 results can be seen in Figure~\ref{fig:twohand_f1} and Figure~\ref{fig:twohand_final_eval_reward}. GPT-6 Astra shows a tendency to improve reward across attempts, whereas the other LLM-based agents exhibit limited or inconsistent reward gains. In contrast, RL methods achieve stronger and more stable reward improvement through iterative training.

    \item[\textbf{Token Consumption.}] The token consumption over ten keyframe-optimization iterations for the two-hand setting on three piano pieces can be seen in Figure~\ref{fig:general-twohand-f1-token-efficiency}. GPT-6 Astra allocates noticeably more tokens to more challenging pieces, whereas Gemini-3.8-Flash and DeepSeek-V4.1-Flash show weaker adaptation of token usage to task difficulty. Claude-Opus-5 maintains the lowest overall token consumption among the evaluated LLM-based agents.

\end{description}

\subsection{Robust RL}
We evaluate robustness over six perturbation targets: action, observation, reward, gravity, contact-friction, and initial hand-pose. Each target is tested under Gaussian, uniform, and shift perturbations with three canonical magnitudes (low, medium, and high), as summarized in Table~\ref{tab:robustness_design}. Signal perturbations are applied at each control step, while physical perturbations are sampled either per episode or per step depending on the setting. Hand-pose perturbations are applied once at reset. This design enables controlled comparison across perturbation type, distribution, magnitude, and frequency.

\begin{table}[H]
\caption{\ourbenchmark robustness axes and canonical low/medium/high levels. Levels are standard deviations for Gaussian noise, half-widths for symmetric uniform noise, and magnitudes for constant shifts. Hand-pose entries report $(y,z)$ offsets.}
\label{tab:robustness_design}
\centering
\small
\setlength{\tabcolsep}{4pt}
\renewcommand{\arraystretch}{1.08}
\begin{tabularx}{\linewidth}{@{}lXllX@{}}
\toprule
Target & Perturbed quantity & Family & Frequency & Low / Medium / High \\
\midrule
Action & Canonical actuator target & G/U/S & Step & $0.05/0.10/0.15$ \\
Observation & Proprioception and piano state & G/U/S & Step & $0.05/0.10/0.15$ \\
Reward & Scalar reward received by the learner & G/U/S & Step & $0.10/0.30/0.50$ \\
Gravity & Vertical acceleration $g_z$ & G/U/S & Episode/step & $0.50/1.00/1.50\,\mathrm{m\,s^{-2}}$ \\
Contact-friction & Fingertip--key sliding coefficient & G/U/S & Episode/step & $0.05/0.10/0.15$ \\
Hand-pose & Per-hand initial $(y,z)$ position & G/U/S & Episode & $(10,5)/(25,10)/(50,20)\,\mathrm{mm}$ \\
\bottomrule
\end{tabularx}
\vspace{-2mm}
\begin{flushleft}
\footnotesize G/U/S denotes Gaussian, uniform, and shift. Signal shifts are fixed across steps; physical shift tasks use an episode-level constant. Step-wise physical variants are provided for Gaussian and uniform perturbations.
\end{flushleft}
\end{table}

\Needspace{15\baselineskip}
\subsection{Safety RL}
\label{detailed safety parameters}
This section details the training parameters used in the safety experiments, summarized in Table~\ref{tab:safety-main-design}.
\begin{table*}[t]
\centering
\small
\setlength{\tabcolsep}{5pt}
\renewcommand{\arraystretch}{1.15}
\caption{Main safety experiments settings. Each task uses PPO, PPO-Lag,
OnCRPO, CPO, and CUP, with three seeds and 5M training steps per run.}
\label{tab:safety-main-design}
\begin{tabularx}{\textwidth}{@{}c X l l r r r@{}}
\toprule
Hands & Song & Cost formulation & Constraint reference
& $T_{\mathrm{nom}}$ & Budget $d$ & Runs \\
\midrule
2 & Für Elise          & Joint-range fraction & Central 50\% & 399 & 19.95 & 15 \\
2 & ClairDeLune       & Joint-range excess   & Central 50\% & 588 & 11.76 & 15 \\
3 & PicturesGreatKiev & Total-power excess   & $P_{\max}=12$ & 720 & 14.40 & 15 \\
3 & PolonaiseOp40No1  & Total-power excess   & $P_{\max}=12$ & 563 & 11.26 & 15 \\
4 & PicturesGreatKiev & Joint-range excess   & Central 50\% & 720 & 14.40 & 15 \\
4 & PicturesGreatKiev & Total-power excess   & $P_{\max}=16$ & 720 & 14.40 & 15 \\

\bottomrule
\end{tabularx}
\par\smallskip
\begin{minipage}{\textwidth}
\footnotesize
$T_{\mathrm{nom}}$ is the episode length in control steps.
Joint-range costs monitor the J2 joints of all five fingers on every hand;
the permitted interval is the central 50\% of each joint's native range.
$P_{\max}$ is the instantaneous total-power reference in simulator units,
not the episode budget. Budgets are fixed as
$d=0.05T_{\mathrm{nom}}$ for event/fraction and
$d=0.02T_{\mathrm{nom}}$ for excess costs.
\end{minipage}
\end{table*}

\begin{table}[H]
\centering
\small
\setlength{\tabcolsep}{8pt}
\renewcommand{\arraystretch}{1.15}
\caption{Baseline algorithms supported by \ourbenchmark across the four evaluation tracks.
\textit{Italic} denotes on-policy methods and \underline{underlined} denotes off-policy methods.}
\label{tab:supported_baselines}

\begin{tabularx}{\textwidth}{@{}lX@{}}
\toprule
Track & Baselines \\
\midrule

Standard RL &
\textit{A2C},
ARS,
\underline{CrossQ},
\underline{DDPG},
\underline{DroQ},
\textit{PPO},
\underline{SAC},
\underline{TD3},
\underline{TQC},
\textit{TRPO}
\\

Robust RL &
\underline{A2P-SAC},
\textit{EPPO},
\underline{OMPO},
\underline{SCPO}
\\

Safe RL &
\textit{PPO-Lag},
\textit{TRPO-Lag},
\textit{RCPO},
\textit{PDO},
\textit{FOCOPS},
\textit{CPO},
\textit{PCPO},
\textit{OnCRPO},
\textit{IPO},
\textit{P3O},
\textit{CUP},
\underline{DDPG-Lag},
\underline{TD3-Lag},
\underline{SAC-Lag}
\\

Multi-Agent RL &
\textit{IPPO},
\textit{MAPPO},
\textit{HAPPO},
\textit{MAT},
\textit{A2PO},
\underline{HATD3},
\underline{FACMAC},
\underline{MACSAC}
\\

\bottomrule
\end{tabularx}

\par\smallskip
\footnotesize
\textit{Italic}: on-policy; \underline{underlined}: off-policy.
ARS is a derivative-free policy-search method.
\end{table}

\end{document}